\pdfoutput=1
\documentclass[11pt]{article}

\usepackage[preprint]{acl}

\usepackage{times}
\usepackage{latexsym}

\usepackage{microtype}
\usepackage{amsmath, amsthm, amssymb, amsbsy}
\usepackage{bm}

\theoremstyle{plain}
\newtheorem{assumption}{Assumption}

\usepackage[T1]{fontenc}
\usepackage[utf8]{inputenc}

\usepackage{microtype}

\usepackage{inconsolata}
\usepackage{amsmath}

\DeclareMathOperator{\E}{\mathbb{E}}
\DeclareMathOperator{\Var}{Var}
\DeclareMathOperator{\Cov}{Cov}

\usepackage{algorithm}
\usepackage{algpseudocode}
\usepackage{algorithmicx}
\usepackage{graphicx}
\usepackage{xspace}
\usepackage{multirow}
\usepackage{booktabs}
\usepackage{amsmath}
\usepackage{enumitem}
\usepackage{amssymb}
\usepackage{makecell}
\usepackage{amsthm}

\newtheorem{theorem}{Theorem}[section]  % numbered within sections
\newtheorem{lemma}[theorem]{Lemma}      % shares numbering with theorem
\newcommand{\ours}{\mbox{{OASIS}}\xspace}
\newcommand{\oursbenchmarkfull}{{\textbf{M}ulti-image \textbf{I}mbalanced \textbf{C}ontinual \textbf{V}isual \textbf{I}nstruction \textbf{T}uning}\xspace}
\newcommand{\oursbenchmark}{\mbox{{MICVIT}}\xspace}
\newcommand{\oursone}{\mbox{{ORIS}}\xspace}
\newcommand{\oursonefull}{{\textbf{O}nline \textbf{R}elative \textbf{I}nformativeness \textbf{S}election}\xspace}
\newcommand{\oursonefullbase}{{Online Relative Informativeness Selection}\xspace}
\newcommand{\ourstwo}{\mbox{{SIREN}}\xspace}
\newcommand{\ourstwofull}{{\textbf{S}imilarity-aware \textbf{I}nformation \textbf{R}edundancy \textbf{E}limi\textbf{N}ation}\xspace}
\newcommand{\ourstwofullbase}{{Similarity-aware Information Redundancy Elimination}\xspace}
\newcommand{\oursfull}{{\textbf{O}nline \textbf{A}daptive \textbf{S}ample selection via \textbf{I}nformative \textbf{S}tatistics}\xspace}
\newcommand{\oursfullbase}{{Online Adaptive Sample selection via Informative Statistics}\xspace}

\newcommand{\eg}{\textit{e.g.}\xspace}
\newcommand{\ie}{\textit{i.e.}\xspace}

\title{\ours: Online Sample Selection for Continual Instruction Tuning}
\author{
    \textbf{Minjae Lee}$^{1,}$\thanks{Equal contribution. $^\dagger$ Corresponding author.} \hspace{.2em}
    \textbf{Minhyuk Seo}$^{2,*}$ \hspace{.2em}
    \textbf{Tingyu Qu}$^2$ \hspace{.2em}
    \textbf{Tinne Tuytelaars}$^{2}$ \hspace{.2em}
    \textbf{Jonghyun Choi}$^{1,\dagger}$ \\
    \\
    $^1$ Seoul National University \quad
    $^2$ KU Leuven \quad \\
    \texttt{\{thkim0305, jonghyunchoi\}@snu.ac.kr} \\
    \texttt{\{minhyuk.seo, tingyu.qu, tinne.tuytelaars\}@kuleuven.be} 
}

\begin{document}
\maketitle

\begin{abstract}
%  To overcome these limitations, we propose \ours, an adaptive online sample selection method that (1) estimates each sample’s informativeness relative to all previously seen data, beyond batch-level constraints, and (2) reduces redundancy among selected samples via iterative score updates. Experiments on various large foundation models show that \ours uses only 25% of the data to match full-data performance and surpasses state-of-the-art sampling methods.
In continual instruction tuning (CIT) scenarios, where new instruction tuning data continuously arrive in an online streaming manner, training delays from large-scale data significantly hinder real-time adaptation. 
Data selection can mitigate this overhead, but existing strategies often rely on pre-trained reference models, which are impractical in CIT setups since future data are unknown.
Recent reference model-free online sample selection methods address this, but typically select a fixed number of samples per batch (\eg, top-$k$), making them vulnerable to distribution shifts where informativeness varies across batches.
To address these limitations, we propose \ours, an adaptive online sample selection approach for CIT that 
(1) selects informative samples by estimating each sample’s informativeness relative to all previously seen data, beyond batch-level constraints, 
and (2) minimizes informative redundancy of selected samples through iterative selection score updates.
Experiments on various large foundation models show that \ours, using only 25\% of the data, achieves comparable performance to full-data training and outperforms the state-of-the-art sampling methods.
\end{abstract}

\section{Introduction}
A key factor in the success of recent large foundation models (LFMs), including LLMs and multi-modal LLMs (MLLMs), is training on large-scale instruction tuning data~\citep{bai2023qwen, chen2023shikra}.
% , with diverse tasks further enhancing generalization~\citep{chen2024feddat}.
While emerging high-quality datasets enable LFMs to better adapt to user preferences and contexts~\citep{maharana2025adaptinfty, lau2024personalized, zhang2024guided}, scaling such datasets also amplifies risks of overfitting and delays in training time~\citep{ma2023understanding, liu2024less}.

\begin{figure}[t]
    \centering   
         \includegraphics[width=\linewidth]{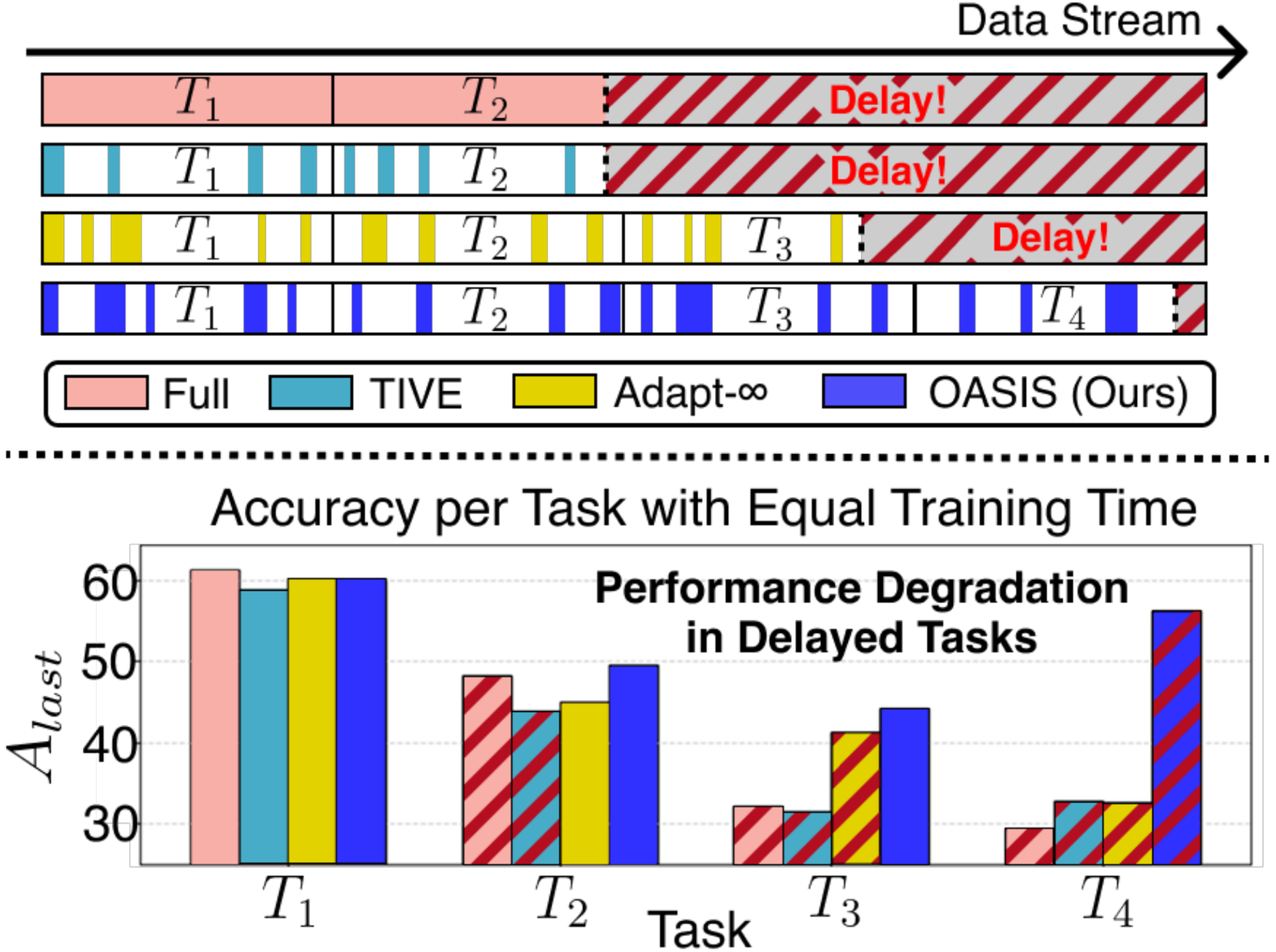}
    \vspace{-1.5em}
    \caption{
    \textbf{Real-time adaptation under equal training time.} Bar width indicates training data volume in the CIT data stream. 
    While 'Full' trains on all data, TIVE~\citep{liu2024less}, Adapt-$\infty$~\citep{maharana2025adaptinfty}, and \ours use 25\% selected data. Under equal training time, 'Full' degrades on newly arrived tasks (\eg, $T_3$, $T_4$), since sequential training on all data provides sufficient time for earlier tasks (\eg, $T_1$, $T_2$) but insufficient time for new ones.
    TIVE and Adapt-$\infty$ achieve only marginal speedup despite using 25\% data, as the backward-pass selection overhead limits real-time adaptation. 
    \ours uses inference-only selection with minimal overhead, enabling efficient sample selection and strong adaptation to new tasks.
    }
    \label{fig:teaser}
    \vspace{-1em}
\end{figure} 

Unlike conventional training paradigms, where LFMs are optimized on static datasets prior to deployment, real-world applications often require adaptation to continuously arriving data streams~\citep{seo2024learning}, \eg, for learning new concepts.
This motivates continual instruction tuning (CIT)~\citep{chen2024coin, maharana2025adaptinfty}, where models adapt to shifting data distributions.
However, training with the massive influx of continuously arriving data in CIT often causes forgetting of prior knowledge~\citep{zhai2023investigating, zhu2024model}, overfitting~\citep{he2024don, ghosh2024closer}, and significant training delays, which hinder the model’s ability to learn from new data promptly (\ie, real-time adaptation)~\citep{koh2021online, caccia2022anytime}.
For example, as shown in Fig.~\ref{fig:teaser}, full-data training 
delays the adaptation to newer tasks ($T_3$ and $T_4$) as the model spends excessive time training on earlier tasks ($T_1$ and $T_2$).

% Unlike conventional training on static datasets prior to deployment, real-world applications require LFMs to adapt to continuously arriving data streams~\citep{seo2024learning}, \eg, for learning new concepts. This setting motivates continual instruction tuning (CIT)~\citep{chen2024coin, maharana2025adaptinfty}, where models adapt to shifting data distributions. However, the massive and ongoing data in CIT often lead to forgetting of prior knowledge~\citep{zhai2023investigating, zhu2024model}, overfitting~\citep{he2024don, ghosh2024closer}, and significant training delays that hinder timely adaptation~\citep{koh2021online, caccia2022anytime}. For instance, as illustrated in Fig.~\ref{fig:teaser}, full-data training slows adaptation to newer tasks ($T_3$, $T_4$) because the model spends excessive time retraining on earlier ones ($T_1$, $T_2$).

Data selection methods~\citep{sorscher2022beyond, lee2024concept, abbas2024effective} mitigate training delays by selecting informative subsets, but most rely on a reference model trained on the entire data in advance~\citep{mindermann2022prioritized, shin2023loss}.
Training a reference model is infeasible in CIT, where new data arrive continuously and future data distributions remain unknown.
While reference model-free methods~\citep{qin2024infobatch, hong2024diversified} eliminate such reliance, their strategy of selecting a fixed number of samples (\eg, top-$k$) from each streaming batch fails to capture inter-batch and intra-batch informativeness.

Specifically, since informative samples are unevenly distributed over time~\citep{seo2025budgeted}, \eg, some batches contain many novel or forgotten samples while others contain few, fixed-size selection per batch often includes uninformative samples while missing critical ones, thereby overlooking inter-batch informativeness variation.
Moreover, in continual data streams, similar instances often recur periodically~\citep{koh2023online}, \eg, Christmas data in winter, swimsuit data in summer.
As a result, many samples within a batch receive similar selection scores~\citep{hong2024diversified}, thus selecting the top-$k$ without considering intra-batch similarity leads to redundancy in the chosen subset.

To address these limitations, we propose \oursfull (\textbf{\ours}), which (i) selects informative samples by estimating each sample’s informativeness relative to all previously encountered data, beyond batch-level constraints, and (ii) reduces the redundancy by considering sample-wise similarity within each batch.
Specifically, \ours maintains online estimates of the global mean and variance of informativeness as batches arrive, and uses these statistics to assess the relative informativeness of each sample.
Moreover, to avoid selecting similar samples within a batch~\citep{hong2024diversified}, \ours iteratively updates the samples' selection scores: once a sample is chosen, the scores of remaining candidate samples in the batch are adjusted according to their shared information with the selected sample.
Note that \ours requires only a forward pass for sample selection, roughly half the cost of a backward pass~\citep{huo2018training}, enabling efficient selection and reducing training delays, thereby facilitating real-time adaptation to newly encountered tasks, as shown in Fig.~\ref{fig:teaser}.

We empirically validate \ours by comparing it with recent sample selection baselines across various LLMs (\eg, LLama-3.1-8B~\citep{grattafiori2024llama} and Qwen3-8B~\citep{yang2025qwen3}) and MLLMs (\eg, LLaVA-1.5-7B~\citep{liu2024improved} and Qwen-VL-2.5-7B~\citep{bai2025qwen2}) on multiple CIT benchmarks.
In particular, \ours incurs only a 1.51\% performance degradation compared to full-data training, while training with only 25\% of samples on \oursbenchmark benchmark.

We summarize our contributions as follows:
\begin{itemize}[nolistsep,noitemsep,topsep=0pt,parsep=0pt,partopsep=0pt]

    \item We propose an adaptive sample selection strategy that selects batch samples based on their informativeness relative to the entire dataset, moving beyond batch-wise selection.

    \item We propose a redundancy reduction strategy that leverages sample-wise similarity to minimize redundancy among selected samples.

    \item By combining these two strategies, our proposed OASIS significantly outperforms baselines in CIT through extensive evaluations.

\end{itemize}

\section{Related Work}
\noindent \textbf{Continual Instruction Tuning.}
Existing instruction tuning methods often focus
on fixed tasks~\citep{zhu2024minigpt, bai2025qwen2}, overlooking continuously emerging instruction tuning data~\citep{maharana2025adaptinfty, guo2025hide}.
To adapt LFMs to dynamically changing data distributions, \emph{continual instruction tuning} (CIT) has been proposed, which aims to learn new tasks while preserving knowledge from previously encountered instruction tuning data~\citep{he2023continual}.
To reflect real-world distribution shifts, various CIT benchmarks have been proposed in both text-only (\eg Long Sequence~\citep{razdaibiedina2023progressive}, TRACE~\citep{wang2023trace}) and multi-modal (\eg COAST~\citep{cao2024continual}, UCIT~\citep{guo2025hide}) domains, along with corresponding strategies (\eg Fwd-Prompt~\citep{zheng2024beyond}, SEFE~\citep{chen2025sefe}).
However, they train LFMs on all data, leading to overfitting~\citep{rice2020overfitting, zhai2023investigating}, high computational costs~\citep{wang2024separable, panos2025efficient}, and training delays~\citep{caccia2022anytime, ghunaim2023real}.
Although aL-SAR~\citep{seo2025budgeted} improves the training efficiency of LFMs through dynamic layer freezing based on batch informativeness, it still relies on the entire dataset, limiting its overall efficiency.

\noindent \textbf{Data Selection.}
Motivated by the observation that not all data contribute equally to learning, recent work explores selecting informative subsets to match full-data performance with lower training cost~\citep{lee2024concept, abbas2024effective, qin2024infobatch}.
Bayesian~\citep{deng2023towards} and RHO-LOSS~\citep{mindermann2022prioritized} enhance training efficiency by training only with a selected subset of data, but both require a reference model pretrained on the full dataset for data selection.
TIVE~\citep{liu2024less} uses smaller reference data, but requires full-layer gradient computation, incurring high computational overhead that offsets any computation savings from data selection.
This reliance on reference models limits their use in CIT, where sequentially arriving data makes full-data pretraining infeasible.
To address this, reference model-free online sample selection methods like GradNorm~\citep{katharopoulos2018not}, InfoBatch~\citep{qin2024infobatch}, and DivBS~\citep{hong2024diversified} have been proposed.
However, they select a fixed number of samples per batch based on difficulty or dissimilarity, limiting adaptability to shifting data distributions, where some batches may contain more informative samples (\eg, newly encountered or forgotten) than others. 
SelfSup~\citep{sorscher2022beyond}, COINCIDE~\citep{lee2024concept}, and DBP~\citep{abbas2024effective} select samples by $K$-means clustering with a sensitive hyperparameter $K$, which is difficult to tune under non-i.i.d. streams due to representation shifts~\citep{ksikazek2025fenec}.
Recently, Adapt-$\infty$~\citep{maharana2025adaptinfty} tackles data selection under shifting data distributions, but it assumes known task boundaries and relies on costly intermediate gradients, limiting its real-world applicability.

\section{Approach}
We first present the problem statement for online sample selection in CIT in Sec.\ref{subsec:prob_statement}. 
We then introduce our method, \oursfull (\textbf{\ours}), in Sec. \ref{subsec:ours}, which consists of two components: \oursonefull (\textbf{\oursone}) in Sec.\ref{subsec:ours_one} and \ourstwofull (\textbf{\ourstwo}) in Sec. \ref{subsec:ours_two}.

\subsection{Problem Statement of Online Sample Selection in CIT} 
\label{subsec:prob_statement}
CIT trains an LFM on a data stream $\mathcal{D}$ comprising a sequence of $T$ tasks, \ie, $\mathcal{D} = {\mathcal{D}_1, \ldots, \mathcal{D}_T}$, where each $\mathcal{D}_i = \{(x^{i}_1, y^{i}_1), (x^{i}_2, y^{i}_2), \dots \}$ denotes the dataset for task $i$.
Note that explicit task boundaries may be absent, and the data distribution across tasks can even be identical.
Reflecting real-world scenarios where data are collected continuously over time~\citep{koh2021online}, online CIT assumes that data arrive as a online stream of samples, denoted as $(x^{i}_1, y^{i}_1), (x^{i}_2, y^{i}_2), \dots$, whereas \textit{offline} CIT provides each task chunk $\mathcal{D}_i$ at once.
At timestep $t$, a batch $\mathcal{B}_{t}$ with batch size $N_\mathcal{B}$ is drawn from $\mathcal{D}$. 
A subset $\mathcal{B}^*_t \subset \mathcal{B}_t$ containing $N_s$ samples ($N_s < N_\mathcal{B}$) is then selected according to a predefined selection ratio, and only this selected subset $\mathcal{B}^*_t$ is used for training.
Given model parameters $\theta$ of an LFM $f$ and loss function $\mathcal{L}$, the objective is:
% A subset $\mathcal{B}^*_t \subset B_t$, consisting of $N_s$ samples ($N_s < N_B$), is then selected according to a predefined selection ratio, and only this selected subset $B_t$ is used for training.
% Given the parameters $\theta$ of an LFM $f$ and a loss function $\mathcal{L}$, the training objective is formulated as:
\begin{equation}
    \min_{\theta} \mathbb{E}_{\mathcal{B}_t \sim \mathcal{D}} \left[ \mathcal{L}\big(f_\theta(\mathcal{B}^*_{t,x}), \mathcal{B}^*_{t,y}\big) \right].
\end{equation}

\begin{figure*}
    \vspace{-.6em}
    \centering   
    \includegraphics[width=\linewidth]{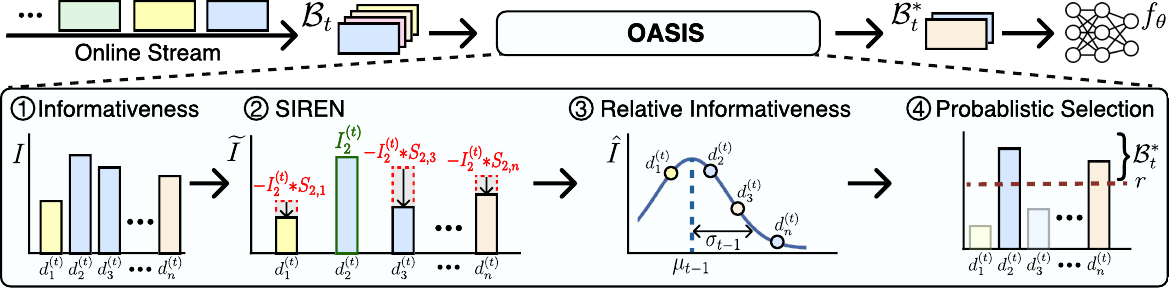}
    \caption{
    \textbf{Overview of our proposed \ours.} 
    For each online batch $\mathcal{B}_t$: 
    (1) \ours first scores the informativeness $I$ for all sample in batch $\mathcal{B}_t=\{d^{(t)}_1, d^{(t)}_2, ...\}$ (Eq.~\ref{eq:select}); 
    (2) It then iteratively reduces redundancy by adjusting $I$ of other samples to $\widetilde{I}$ based on their gradient similarity $S_{i,j}$ to the most informative sample (here, $d^{(t)}_2$). (Sec.~\ref{subsec:ours_two});
    (3) \ours computes relative informativeness $\hat{I}$ by normalizing the updated informativeness $\widetilde{I}$ using EMA $\mu_t$ and EMV $\sigma_t$ (Eq.~\ref{eq:normalize}); 
    (4) Finally, \ours computes selection probability $P_S$ and selects samples exceeding a uniformly drawn threshold $r$, resulting in a selected subset $\mathcal{B}^*_t \subset \mathcal{B}_t$ (Eq.~\ref{eq:prob_selection}).
    Model $f_{\theta}$ is then trained using only $\mathcal{B}^*_t$.
    % For each online batch, (1) \ours first scores the informativeness $I$ of each sample. (Sec.~\ref{subsec:ours_one})
    % (2) It then iteratively reduces redundancy by adjusting $I$ of other samples based on their gradient similarity $S_{i,j}$ to the most informative sample $d_i$. (Sec.~\ref{subsec:ours_two})
    % (3) Next, \ours computes the relative informativeness $\hat{I}$ by normalizing based on EMA $\mu_t$ and EMV $\sigma_t$ of $I$ the adjusted $I$. (Sec.~\ref{subsec:ours_one})
    % (4) Finally, \ours computes each sample's selection probability and picks those exceeding a uniformly drawn threshold $r$. (Sec.~\ref{subsec:ours_one})
    }
    \label{fig:overview}
    \vspace{-.5em}
\end{figure*} 

\vspace{-0.5em}
\subsection{Online Adaptive Sample Selection via Informative Statistics}
\label{subsec:ours}

Existing online sample selection methods typically select a fixed number of samples per batch, \eg, top-$N_S$ selection scores, which struggles under distribution shifts (\eg, CIT), due to: (i) varying numbers of informative samples per batch over time, and (ii) redundancy among co-occurring similar samples within a batch~\citep{hong2024diversified}.

To address these limitations, we propose two strategies: (i) \oursonefullbase (\oursone), which selects samples based on their relative informativeness across the entire data distribution, rather than intra-batch ranking; and (ii) \ourstwofullbase (\ourstwo), which mitigates redundancy by updating, at each selection step, the selection scores of remaining samples within a batch according to their similarity with already selected ones.

% which reduces redundancy by adjusting sample informativeness based on sample-wise gradient similarity.

Specifically, for each training batch, \oursone computes sample-wise informativeness using Fisher Information (FI) and estimates relative informativeness within the overall FI distribution.
During this calculation, \ourstwo adjusts these scores using sample-wise gradient similarity to account for the impact of training one sample on others. 
Integrating \oursone and \ourstwo, we call our method \oursfullbase (\ours).
We provide an overview of \ours in Fig.~\ref{fig:overview}, and pseudocode in Sec.~\ref{sec:appx:pseudocode}.

% \subsection{\oursfull}

\subsection{Online Relative Informativeness Selection}
\label{subsec:ours_one}

% We score each sample by a Fisher Information–motivated measure that estimates its potential contribution to learning \citep{deng2023uncertainty}.
% Computing full FI is costly (requires forward–backward over all layers), so we adopt an efficient approximation: (i) evaluate gradients only at the last layer $\theta_L$—earlier-layer gradients scale by the chain rule \citep{koh2023online}; and (ii) use a first-order, diagonal FI approximation that avoids Hessian computation \citep{kirkpatrick2017overcoming,soen2021variance}.
% For the $i$-th sample $d^{(t)}i=(x^{(t)}i,y^{(t)}i)$ in batch $B_t$, its informativeness is
% \begin{equation}
% \label{eq:select}
% I_i^{(t)}=\mathrm{tr}!\left[!\big(\nabla{\theta_L} L(d_i^{(t)})\big)\big(\nabla{\theta_L} L(d_i^{(t)})\big)^{!\top}\right]
% = \big|\nabla{\theta_L} L(d_i^{(t)})\big|2^2,
% \end{equation}
% where $L(d_i^{(t)})=\ell(f\theta(x_i^{(t)}),y_i^{(t)})$ and $\ell$ is the loss.
% Unlike methods that use middle or all layers \citep{maharana2025adaptinfty,liu2024less}—with cost close to full-data training—\ours{} requires only last-layer gradients, yielding near inference-only overhead.

\noindent \textbf{Informativeness $I$.}
To select informative samples, we first define the sample-wise informativeness $I$ based on Fisher Information (FI), which measures a model’s potential information gain from each example \citep{deng2023uncertainty}.
Computing full FI is costly, since it requires forward and backward passes across all layers, which would negate the efficiency gains of training on only a selected subset.
We therefore adopt an efficient approximation: (i) computing gradients only at the last layer $\theta_L$, since gradients of preceding layers are proportional to it by the chain rule~\citep{koh2023online}, and (ii) using a first-order diagonal FI approximation that avoids Hessian computation \citep{kirkpatrick2017overcoming,soen2021variance}.
Formally, we define the informativeness of the $i_\text{th}$ sample $d^{(t)}_i = (x^{(t)}_i, y^{(t)}_i)$ in batch $\mathcal{B}_t$ as:
\begin{equation}
    \label{eq:select}
    \!I_i^{(t)}\! =\! \mathrm{tr} \left[ \left(\nabla_{\theta_L} L(d^{(t)}_i) \right)  \cdot \left(\nabla_{\theta_L} L(d^{(t)}_i) \right)^{\top} \right],\!
\end{equation}
where $L(d^{(t)}_i) = \ell\big(f_\theta(x_i^{(t)}), y_i^{(t)}\big)$, $\ell$ is the loss function, and $\mathrm{tr}(\cdot)$ denotes the trace operator.
While existing methods use gradients from middle~\citep{maharana2025adaptinfty} or all layers~\citep{liu2024less} for selection, incurring costs close to full-data training, \oursone requires only last-layer gradients, yielding near inference-only cost.

% Cross-batch comparison would require recomputing 
% ��
% I for all past samples, which is expensive and often infeasible in online CIT where past data may be discarded \citep{maharana2025adaptinfty}. Instead of storing all 
% ��
% I values, we maintain a running mean of 
% ��
% I over seen samples and assess each new sample by its deviation from this mean, yielding a relative informativeness 

\noindent \textbf{Relative Informativeness $\hat{I}$.}
However, we cannot compare $I$ across batches, as $\theta$ in Eq.~\ref{eq:select} changes over time due to model updates at each batch iteration.
In other words, cross-batch comparison of $I$ would require re-forwarding all previously encountered samples with the latest $\theta$, which is computationally prohibitive and often infeasible in online CIT scenarios where previous samples may be discarded~\citep{maharana2025adaptinfty}.
To avoid these issues, instead of tracking all samples' $I$, we maintain a running mean of $I$ over seen samples and evaluate each new sample by its deviation from this mean, yielding a relative informativeness over all previously observed data.
% we estimate each sample's relative position with respect to the average $I$ of encountered samples.

To be specific, inspired by the exponential decay model of forgetting~\citep{shin2020new, mahto2020multi, chien2021slower}, we maintain exponential moving average (EMA) and variance (EMV) of $I$, which exponentially decay the influence of previous statistics.
Formally, at each timestep $t$, given batch $\mathcal{B}_t$ of size $N_\mathcal{B}$ and decay factor $\beta$, we update EMA $\mu_t$ and EMV $\sigma_t$ of $I$ as: 
\begin{equation}
\label{eq:EMA_EMV}
\begin{aligned}
    \mu_{t} &= \beta \bar{I}^{(t)} + (1-\beta)\mu_{t-1}, \\
    \sigma_{t} &= \beta (\bar{I}^{(t)} - \mu_{t-1})^2 + (1-\beta)\sigma_{t-1}.
\end{aligned}
\end{equation}
where $\bar{I}^{(t)} = \frac{1}{N_\mathcal{B}}\sum_{i=1}^{N_\mathcal{B}} I^{(t)}_i$ denotes average $I^{(t)}$ of samples in $\mathcal{B}_t$.

Using these statistics, we estimate each sample's relative informativeness.
% over all previously observed data.
By Theorem~\ref{thm:main_normal}, normalization with the EMA $\mu_{t-1}$ and EMV $\sigma_{t-1}$ (updated up to step $t-1$) yields a standard normal distribution, enabling deviation measurement of each sample via $Z$-score normalization (see Sec.~\ref{sec:appx:proof_thm1} for proof).
Formally, we define relative informativeness of the $i_\text{th}$ sample with informativeness $I^{(t)}_i$ in batch $\mathcal{B}_t$ at time step $t$ as:
\begin{equation}
    \label{eq:normalize}
    \hat{I}^{(t)}_i = \frac{I^{(t)}_i - \mu_{t-1}}{\sigma_{t-1}}.
\end{equation}

% We use $\mu{t-1},\sigma_{t-1}$ rather than $\mu_t,\sigma_t$ to avoid self-normalization bias~\citep{jing2003self, delapena2009self}, ensuring independence from the current sample.
Note that we normalize with $\mu_{t-1}$ and $\sigma_{t-1}$ rather than $\mu_t$ and $\sigma_t$, since the latter already incorporate $I^{(t)}_i$ and thus induce dependence between numerator and denominator (\ie, self-normalization bias~\citep{jing2003self, delapena2009self}). 
% Using $\mu_{t-1},\sigma_{t-1}$ ensures independence from the current sample by relying only on past data.

% As empirical evidence shows that $I$ of observed data approximates a normal distribution (Fig.~\ref{fig:qqplot}), we apply Z-score normalization to $I$ to quantify a sample's deviation from the overall $I$ distribution.

% \begin{theorem}[One-step normalized FI scores are approximately standard normal]
% \label{thm:main_normal}
% Let $I^{(t)}_i$ be defined by Eq.~\ref{eq:select} for samples in $B_t$, and let $(\mu_{t-1},\sigma^2_{t-1})$ be the EMA and EMV from Eq.~\ref{eq:EMA_EMV} computed from batches prior to $t$.
% Assume local stationarity and local weak dependence of $\{I^{(t)}_i\}$ (Assumptions~\ref{asmp:local_stationarity}, \ref{asmp:local_weakdep}) with uniformly bounded second moments.
% Then the one-step standardized statistic
% \[
% \hat{I}^{(t)}_i\;=\;\frac{I^{(t)}_i-\mu_{t-1}}{\sigma_{t-1}}
% \]
% satisfies $\hat{I}^{(t)}_i\approx \mathcal N(0,1)$.
% \end{theorem}

\begin{theorem}
\label{thm:main_normal}
Let $I^{(t)}_i$ be defined by Eq.~\ref{eq:select} for samples in $\mathcal{B}_t$, and let $(\mu_{t-1},\sigma^2_{t-1})$ be the EMA and EMV from Eq.~\ref{eq:EMA_EMV} computed from batches prior to $t$.
Assume local stationarity and local weak dependence of $\{I^{(t)}_i\}$ with uniformly bounded second moments.
Then
\begin{equation}
    \hat{I}^{(t)}_i \;=\; \frac{I^{(t)}_i-\mu_{t-1}}{\sigma_{t-1}}
    \;\approx\; \mathcal N(0,1).
\end{equation}
\end{theorem}

\noindent \textbf{Probabilistic Selection.}
We then determine whether to select each sample based on $\hat{I}^{(t)}$.
Inspired by coreset methods~\citep{bang2021rainbow, seo2024just}, which select both informative and easy-to-learn samples, we adopt probabilistic selection based on $\hat{I}$ rather than deterministically selecting those above a fixed threshold $I_T$ (set by the target selection ratio).
Specifically, we apply Sigmoid to $\hat{I} - I_{T}$, and perform Bernoulli sampling, allowing samples below the threshold to be included with nonzero probability.
Formally, with $r \sim \mathcal{U}(0,1)$, where $\mathcal{U}(0,1)$ denotes the uniform distribution over the interval $[0,1]$, we define the selected samples set $\mathcal{B}^*_t$ from batch $\mathcal{B}_t$ as:
\begin{equation}
    \label{eq:prob_selection}
    \mathcal{B}^*_t =  \{d^{(t)}_i \in \mathcal{B}_t \mid \mathrm{sigmoid}(\hat{I}^{(t)}_i - I_{T}) > r\}.
\end{equation}

% Unlike existing online selection methods that require knowledge of the total sample count to fix per-batch quotas, \oursone uses only the selection ratio to determine $I_T$, enabling operation on unbounded data streams. Details on computing $I_T$ are provided in Sec.~\ref{sec:appx:threshold}.

Note that existing online selection methods require knowledge of the total sample count to fix per-batch quotas. 
In contrast, \oursone uses only the selection ratio to determine $I_T$, enabling operation on endless real-world data streams.
We provide details on determining $I_{T}$ in Sec.~\ref{sec:appx:threshold}.

% ema 현재 강조
% \begin{figure}[t!]
%     % \vspace{-.6em}
%     \centering   
%     \includegraphics[width=\linewidth]{figures/qqplot.pdf}
%     \vspace{-1.5em}
%     \caption{
%     \textbf{QQ Plot of informativeness $I$.}
%     We measure $I$ on 500 randomly selected samples at each training timestep. 
%     Blue dots denote samples, and the red line denotes the reference normal distribution.
%     Their close alignment suggests that $I$ is approximately normally distributed.
%     }
%     \label{fig:qqplot}
%     \vspace{-1em}
% \end{figure} 

\subsection{Similarity-Aware Information Redundancy Elimination}
\label{subsec:ours_two}

% In real-world data streams, similar task instances often appear consecutively. Selecting samples within $B_t$ using \oursone without accounting for intra-batch similarity can therefore lead to redundant choices. 
% To mitigate this, we adjust $I$ based on sample-wise similarity. 
% Ideally, one could iteratively train on the most informative sample in a batch and then recompute $I$ for the remaining samples, as training on similar data reduces their informativeness~\citep{hekimoglu2023active}. 
% However, this requires repeated $I$ computation and is computationally prohibitive for batch-wise training.

In real-world data streams, similar task samples often arrive consecutively. Selecting samples within $\mathcal{B}_t$ using \oursone without considering intra-batch similarity can thus lead to redundancy. To mitigate this, we adjust $I$ based on sample-wise similarity. 
Ideally, one would iteratively select the most informative unchosen sample, train the model, and then recompute $I$ for the remaining candidates using the updated model, thereby accounting for the reduced informativeness of samples similar to those already trained on~\citep{hekimoglu2023active}. 
However, this procedure incurs prohibitive computational cost due to repeated $I$ computation, rendering batch-wise training impractical.

% Instead, following~\citet{du2018adapting, seo2025budgeted}, we approximate this process by leveraging gradient similarity, which correlates with informational overlap. When a sample is selected, we reduce the $I$ of remaining samples proportionally to their cosine similarity with its gradient:
% \begin{equation}
% \widetilde{I}i^{(t)} = I_i^{(t)} - \sum{h \in H} \cos(g_i, g_h), I_h,
% \end{equation}
% where $g_i = \nabla\ell_{\theta_L}(f(x_i), y_i)$ and $H$ is the set of already selected samples. We iteratively update $\widetilde{I}$ within the batch to obtain a more diverse and informative subset, as summarized in Algorithm~\ref{algo:pseudo}.

% iteratively update each sample’s $I$ to simulate its change, assuming the model has already been trained on previously identified high-informative samples.
Instead of re-forwarding to recompute $I$, we approximate its change induced by selecting other samples.
Inspired by~\citet{du2018adapting, seo2025budgeted}, which show that gradient alignment correlates with informational overlap, when a sample is selected, we multiply its $I$ by its gradient similarity with each remaining sample and subtract the result from their $I$ values, thereby reducing the informativeness of samples that overlap with the selected one.
Formally, given a batch $\mathcal{B}_t = \{d^{(t)}_i\}_{i=1}^{N_\mathcal{B}}$, we first calculate $I$ for each sample and sort them in descending order, \ie, $I^{(t)}_1 > \cdots > I^{(t)}_{N_\mathcal{B}}$.
We then add the most informative sample (\ie, $d_1$) to the set $H$, the set of samples that is assumed to have been trained on.
For each remaining sample $d^{(t)}_i \in \mathcal{B}_t \setminus H$, we define updated informativeness $\widetilde{I}_i^{(t)}$ as:
\begin{equation}
\widetilde{I}_i^{(t)} = I_i^{(t)} - \sum_{h \in H} \text{cos}(g_i, g_h) \cdot I_h,
\end{equation}
where $g_i = \nabla\ell_{\theta_L}(f(x_i), y_i)$ refers to the last-layer gradient of sample $d_i=(x_i, y_i)$.
We iteratively add $d_1, \cdots, d_{N_\mathcal{B}}$ to $H$ and repeat the process iteratively, as shown in Algorithm~\ref{algo:pseudo}.

However, when $|H| > 1$, overlapping similarities between samples in $H$ cause duplicate subtractions. 
% To address this, we incorporate higher-order redundancy using the inclusion-exclusion principle as:
We resolve this using the inclusion-exclusion principle to capture higher-order redundancy as:
\begin{equation}
\begin{split}
\widetilde{I}_i^{(t)} &= I_i^{(t)} - \sum_{h \in H} \text{cos}(g_i, g_h)\ \cdot I_h \\
&+ \sum_{\substack{U \subseteq H \\ |U| \geq 2}} (-1)^{|U|} \text{cos}(g_i, \bar{g}_U) \cdot \bar{I}_{U},
\end{split}
\end{equation}
where $U$ denotes all possible non-empty subsets of $H$ with $|U|\geq 2$, $\bar{g}_U=\frac{1}{|U|}\sum_{u \in U}g_u$ and $\bar{I}_{U} =\frac{1}{|U|} \sum_{u \in U} I_u$ denote the average gradient and average $I$ over subset $U \subseteq H$, respectively.

We use $\widetilde{I}^{(t)}$to calculate relative informativeness $\hat{I}^{(t)}$ instead of directly using $I^{(t)}$, as shown in Fig.~\ref{fig:overview}.
By applying \ourstwo before computing $\hat{I}^{(t)}$ in \oursone, we effectively reduce redundancy without requiring any re-forwarding process.

\section{Experiments}
\label{sec:experiments}

\subsection{Experimental Setup}
\label{sec:experimental_setup}
In all experiments, we report the mean and standard deviation of the results from three different seeds.
Moreover, we conduct Welch’s t-test with a significance level of 0.05. 
The highest performance is highlighted in bold, and results not significantly different from the best are \underline{underlined}.

\noindent \textbf{Models.}
For MLLMs, we use LLaVA-1.5~\citep{liu2024improved} and Qwen-VL-2.5~\citep{bai2025qwen2} as our MLLMs. 
In the main paper, we focus on LLaVA-1.5-7B and Qwen-VL-2.5-7B, while we provide experiments with other model sizes of LLaVAs (1B, 3B, and 13B) and Qwen-VLs (0.5B), in Sec.~\ref{sec:appx:scale_exp}.
During training, we update only the LoRA adapters~\citep{hu2022lora}, keeping the LLM frozen for training efficiency~\citep{ye2023mplug}.
For LLMs, we use Llama-3.1-8B~\citep{grattafiori2024llama} and Qwen3-8B~\citep{yang2025qwen3}.

\noindent \textbf{Baselines.}
% We compare our proposed \ours with 
We compare \ours with
recent state-of-the-art sample selection methods, including GradNorm~\citep{katharopoulos2018not}, Self-Sup~\citep{sorscher2022beyond}, COINCIDE~\citep{lee2024concept}, DBP~\citep{abbas2024effective}, InfoBatch~\citep{qin2024infobatch}, DivBS~\citep{hong2024diversified}, TIVE~\citep{liu2024less}, Adapt-$\infty$~\citep{maharana2025adaptinfty}, and Random selection, which, despite its simplicity, often surpasses SOTA selection methods as noted in~\citet{gupta2023data, zheng2023coveragecentric, maharana2025adaptinfty}.
While baselines select a fixed number of samples per batch, \ours dynamically selects samples probabilistically.
For fair comparisons, we ensure \ours uses a comparable or fewer total samples (see Sec.~\ref{sec:appx:num_samples}.
See Sec.~\ref{sec:appx:baselines} for the baselines' details.

% \vspace{-0.5em}
% \paragraph{Metrics.}
\noindent \textbf{Metrics.}
We report $A_{last}$, the accuracy measured at the end of training, and $A_{avg}$, the average accuracy measured at each task boundary.

\noindent \textbf{Benchmarks.}
We evaluate on a range of CIT benchmarks, including text-only benchmarks (Long Sequence~\citep{razdaibiedina2023progressive}, TRACE~\citep{wang2023trace}) and multi-modal benchmarks (COAST~\citep{cao2024continual}, Adapt~\citep{maharana2025adaptinfty}).
However, COAST assumes balanced task sizes (\ie, 20k samples per task), which contradicts real-world data imbalance, while Adapt includes datasets containing COCO~\citep{lin2014microsoft} images that overlap with LLaVA’s instruction-tuning data.
To address these limitations, we introduce \oursbenchmarkfull (\textbf{\oursbenchmark}), a new benchmark that removes such overlaps while preserving naturally imbalanced task distributions.
To better reflect real-world complexity, \oursbenchmark primarily adopts multi-image datasets.

More details on implementation and benchmarks can be found in Sec.~\ref{sec:appx:implementation_details} and~\ref{sec:appx:benchmark_details}, respectively.

\subsection{Quantitative Analysis}
\label{sec:quanti_main}

\noindent \textbf{Comparison across Various Sample Selection Ratios.}
We compare \ours with the baselines by training LLaVA-1.5-7B on \oursbenchmark under various selection ratios (\ie, 6.25\%, 12.5\%, and 25.0\%).
As shown in Tab. \ref{tab:main_llava_ours}, \ours significantly outperforms baselines, with the largest gap at the lowest sample selection ratio (\ie, 6.25\%).
Since each sample has a greater impact at smaller coreset sizes~\citep{zheng2023coveragecentric, jafari2024power}, this highlights \ours’s strength in identifying informative data.
Moreover, at the 25.0\% ratio, \ours nearly matches full-data training, showing only a 1–2\% drop in $A_{last}$.
Finally, gains in both $A_{last}$ and $A_{avg}$ indicate that \ours not only boosts final accuracy but also accelerates convergence by prioritizing informative samples.

Methods using K-means clustering (\ie, Self-Sup, COINCIDE, DBP, TIVE, and Adapt-$\infty$) often perform worse than random selection due to imbalanced clustering under imbalanced data distributions, as detailed in Sec.~\ref{sec:appx:selection_distribution}.

% We evaluate \ours against baselines by training LLaVA-1.5-7B on \oursbenchmark under different sample selection ratios (6.25%, 12.5%, 25.0%), as shown in Tab.~\ref{tab:main_llava_ours}. \ours consistently outperforms all baselines, with the largest margin at 6.25%, where each sample has greater influence~\citep{zheng2023coveragecentric, jafari2024power}, underscoring its ability to identify informative data. At 25.0%, \ours nearly matches full-data training, showing only a 1–2% drop in $A_{last}$. Gains in both $A_{last}$ and $A_{avg}$ indicate that \ours not only boosts final accuracy but also accelerates convergence by prioritizing informative samples.

\noindent \textbf{Comparison across Various Benchmarks.}
We also compare \ours with baselines using a 6.25\% selection ratio on existing CIT benchmarks, namely COAST, Adapt, Long Sequence, and TRACE (Tab.~\ref{tab:main_llava_coast_adapt_625}).
As shown in the table, \ours consistently outperforms the baselines across multiple CIT benchmarks, demonstrating its robustness and generalizability.
We provide additional results with various selection ratios in Sec.~\ref{sec:appx:llava_benchmark_results}

\noindent \textbf{Comparison with Qwen-VL.} 
Beyond LLaVA-1.5-7B, we further test \ours on Qwen-VL-2.5-7B.
As shown in Tab.~\ref{tab:main_qwen_ours}, \ours consistently outperforms baselines, demonstrating its model-agnostic generalizability.
See Sec.~\ref{sec:appx:qwen_results} for results with different selection ratios and benchmarks on Qwen-VL.

\begin{table}[h]
    % \vspace{-0.1em}
  % \label{tab:headings}
  \centering
  \resizebox{\linewidth}{!}{
    \begin{tabular}{lcc}
    \toprule
    \multirow{1}{*}{Method} 
    & $A_{avg} \ \uparrow$ & $A_{last} \ \uparrow$ \\ 
    
    \cmidrule(lr){1-1} \cmidrule(lr){2-2} \cmidrule(lr){3-3} 
    Full-Data Training & 
    72.31$\pm$0.42 & 78.18$\pm$0.75 \\

    \cmidrule(lr){1-1} \cmidrule(lr){2-2} \cmidrule(lr){3-3}  

    Random & 
    67.44$\pm$0.12 & 74.12$\pm$0.42 \\

    GradNorm {\footnotesize \color{blue}{(ICML 2018)}} & 
    67.91$\pm$0.55 & 73.39$\pm$0.30 \\

    Self-Sup {\footnotesize \color{blue}{(NeurIPS 2022)}} & 
    64.34$\pm$0.49 & 70.71$\pm$0.56 \\
    
    COINCIDE {\footnotesize \color{blue}{(EMNLP 2024)}} & 
    65.12$\pm$0.38 & 71.04$\pm$0.72 \\
    
    DBP {\footnotesize \color{blue}{(ICLR 2024)}} & 
    62.42$\pm$0.42 & 72.24$\pm$0.61 \\

    InfoBatch {\footnotesize \color{blue}{(ICLR 2024)}} & 
    65.69$\pm$0.30 & 73.51$\pm$0.45 \\

    DivBS {\footnotesize \color{blue}{(ICML 2024)}} & 
    66.18$\pm$0.74 & \underline{75.27$\pm$0.42} \\

    TIVE {\footnotesize \color{blue}{(arXiv:2403)}} & 
    64.64$\pm$0.59 & 72.16$\pm$0.81 \\

    Adapt-$\infty$ {\footnotesize \textcolor{blue}{(ICLR 2025)}} &
    65.95$\pm$0.61 & 72.18$\pm$0.16 \\

    \cmidrule(lr){1-1} \cmidrule(lr){2-2} \cmidrule(lr){3-3} 
    
    \ours (\textbf{Ours}) & 
    \textbf{70.23}$\pm$\textbf{0.27} & \textbf{76.4}1$\pm$\textbf{0.41} \\
    \bottomrule
    \end{tabular}
    }
    \vspace{-0.5em}
    \caption{
    \textbf{Quantitative comparison with Qwen-VL-2.5-7B on \oursbenchmark at a 25.0\% selection ratio.} Bold indicates the highest performance; underlined results are within the 0.05 t-test significance level.
    }
  \label{tab:main_qwen_ours}
  % \vspace{-.5em}
\end{table}

\begin{table*}[t]
  % \label{tab:headings}
  \vspace{-0.5em}
  \centering
  \resizebox{\linewidth}{!}{
    \begin{tabular}{lcccccc}
    \toprule
    \multirow{3}{*}{Method} & \multicolumn{6}{c}{Selection Ratio (\%)} \\     \cmidrule(lr){2-7} 
    & \multicolumn{2}{c}{6.25} & \multicolumn{2}{c}{12.5} & \multicolumn{2}{c}{25.0} \\
    
    & $A_{avg} \ \uparrow$ & $A_{last} \ \uparrow$ & $A_{avg} \ \uparrow$ & $A_{last} \ \uparrow$ & $A_{avg} \ \uparrow$ & $A_{last} \ \uparrow$ \\ 
    
    % \cmidrule(lr){1-1} \cmidrule(lr){2-3} \cmidrule(lr){4-5} \cmidrule(lr){6-7} 

    % Full-Training & 
    % \multicolumn{6}{c}{ $\pm$ } \\
    % \multicolumn{6}{c}{71.80$\pm$0.44 79.66$\pm$0.43 } \\
    \cmidrule(lr){1-1} \cmidrule(lr){2-3} \cmidrule(lr){4-5} \cmidrule(lr){6-7} 
    Full-Data Training & 
    71.80$\pm$0.44 & 79.66$\pm$0.43 & 
    71.80$\pm$0.44 & 79.66$\pm$0.43 & 71.80$\pm$0.44 & 79.66$\pm$0.43 \\
    
    \cmidrule(lr){1-1} \cmidrule(lr){2-3} \cmidrule(lr){4-5} \cmidrule(lr){6-7} 

    Random & 
    61.15$\pm$0.34 & 67.29$\pm$0.61 & 64.50$\pm$0.22 & 
    71.33$\pm$0.47 & 65.48$\pm$0.73 & 73.84$\pm$0.45 \\

    GradNorm {\footnotesize \color{blue}{(ICML 2018)}} & 
    61.23$\pm$0.83 & 66.33$\pm$0.73 & 63.68$\pm$1.39 & 
    70.51$\pm$0.10 & \underline{65.81$\pm$1.37} & 72.03$\pm$0.94 \\

    Self-Sup {\footnotesize \color{blue}{(NeurIPS 2022)}} & 
    58.05$\pm$0.77 & 64.61$\pm$1.16 & 62.84$\pm$0.73 & 
    69.48$\pm$0.62 & 64.39$\pm$1.26 & 71.81$\pm$0.13 \\
    
    COINCIDE {\footnotesize \color{blue}{(EMNLP 2024)}} & 
    59.28$\pm$1.45 & 64.83$\pm$0.45 & 62.33$\pm$0.92 & 
    69.50$\pm$0.39 & 64.92$\pm$0.81 & 72.25$\pm$1.17 \\
    
    DBP {\footnotesize \color{blue}{(ICLR 2024)}} & 
    57.14$\pm$0.51 & 63.24$\pm$0.82 & 60.64$\pm$1.06 & 
    67.85$\pm$1.19 & 62.14$\pm$1.42 & 70.13$\pm$1.39 \\

    InfoBatch {\footnotesize \color{blue}{(ICLR 2024)}} & 
    60.82$\pm$0.75 & 68.88$\pm$1.09 & 64.87$\pm$0.68 & 
    73.63$\pm$0.28 & \underline{66.41$\pm$1.36} & 74.70$\pm$0.52 \\

    DivBS {\footnotesize \color{blue}{(ICML 2024)}} & 
    61.07$\pm$0.58 & 69.06$\pm$0.54 & 65.31$\pm$0.41 & 
    \underline{73.76$\pm$0.85} & 64.22$\pm$0.71 & \underline{75.38$\pm$1.25} \\

    TIVE {\footnotesize \color{blue}{(arXiv:2403)}} & 
    58.75$\pm$1.20 & 64.19$\pm$1.40 & 61.27$\pm$0.59 & 
    68.38$\pm$0.41 & 64.29$\pm$1.02 & 72.76$\pm$0.83 \\

    Adapt-$\infty$ {\footnotesize \textcolor{blue}{(ICLR 2025)}} &
    59.13$\pm$0.32 & 64.37$\pm$0.85 & 61.53$\pm$0.84 & 
    69.82$\pm$0.99 & 65.50$\pm$0.52 & 73.57$\pm$1.22 \\

    \cmidrule(lr){1-1} \cmidrule(lr){2-3} \cmidrule(lr){4-5} \cmidrule(lr){6-7} 
    
    \ours (\textbf{Ours}) & 
    \textbf{64.39}$\pm$\textbf{0.58} & \textbf{71.76}$\pm$\textbf{0.72} & \textbf{66.89}$\pm$\textbf{0.68} & 
    \textbf{75.60}$\pm$\textbf{0.26} & \textbf{68.84}$\pm$\textbf{0.37} & \textbf{77.95}$\pm$\textbf{0.93} \\
    
    \bottomrule
    \end{tabular}
    }
    \vspace{-0.5em}
    \caption{
    \textbf{Quantitative comparison between sample selection methods on \oursbenchmark benchmark.}
    % We use LLaVA-1.5-7B as the MLLM.
    % Bold indicates the highest performance, and underlined results are within the 0.05 significance level of the t-test.
    % `Full Data Training' denotes training the MLLM on all incoming data without data selection.
    LLaVA-1.5-7B is used as the MLLM.
    Bold indicates the highest performance; underlined results are within the 0.05 t-test significance level. 
    'Full-Data Training' uses all data without selection.
    % \ours outperforms the baselines across various selection ratios.
    }
  \label{tab:main_llava_ours}
  \vspace{-.6em}
\end{table*}

\begin{table*}[t]
  % \label{tab:headings}
  % \vspace{-0.5em}
  \centering
  \resizebox{\linewidth}{!}{
    \begin{tabular}{lcccccccc}
    \toprule
    \multirow{3}{*}{Method} & \multicolumn{4}{c}{MLLM Benchmark} & \multicolumn{4}{c}{LLM Benchmark} \\     \cmidrule(lr){2-5} \cmidrule(lr){6-9} 
    & \multicolumn{2}{c}{COAST} & \multicolumn{2}{c}{Adapt} & \multicolumn{2}{c}{Long Sequence} & \multicolumn{2}{c}{TRACE} \\
    
    & $A_{avg} \ \uparrow$ & $A_{last} \ \uparrow$ & $A_{avg} \ \uparrow$ & $A_{last} \ \uparrow$ & $A_{avg} \ \uparrow$ & $A_{last} \ \uparrow$ & $A_{avg} \ \uparrow$ & $A_{last} \ \uparrow$ \\ 
    
    % \cmidrule(lr){1-1} \cmidrule(lr){2-3} \cmidrule(lr){4-5} \cmidrule(lr){6-7} 

    % Full-Training &
    % \multicolumn{6}{c}{ $\pm$ } \\
    % \multicolumn{6}{c}{ 31.56$\pm$1.42 39.06$\pm$0.55 } \\
    \cmidrule(lr){1-1} \cmidrule(lr){2-3} \cmidrule(lr){4-5} \cmidrule(lr){6-7} \cmidrule(lr){8-9} 
    Full-Data Training & 
    31.56$\pm$1.42 & 39.06$\pm$0.55 & 55.89$\pm$0.20 & 54.26$\pm$0.23 & 85.83$\pm$0.48 & 82.83$\pm$0.47 & 50.84$\pm$0.70 & 57.73$\pm$0.54 \\
    
    \cmidrule(lr){1-1} \cmidrule(lr){2-3} \cmidrule(lr){4-5} \cmidrule(lr){6-7} \cmidrule(lr){8-9} 

    Random & 
    23.57$\pm$0.17 & 30.80$\pm$0.30 & 47.64$\pm$0.21 & 40.26$\pm$0.71 & 
    70.90$\pm$0.13 & 68.02$\pm$0.18 & 43.44$\pm$0.55 & 52.27$\pm$0.03 \\

    GradNorm & 
    21.91$\pm$0.47 & 28.08$\pm$0.49 & 44.92$\pm$0.06 & 36.78$\pm$0.52 &
    70.70$\pm$0.05 & 68.36$\pm$0.75 & 41.39$\pm$0.86 & 50.29$\pm$0.69 \\

    Self-Sup & 
    22.16$\pm$0.27 & 30.09$\pm$0.51 & 43.85$\pm$0.75 & 34.49$\pm$1.37 & 
    67.88$\pm$0.28 & 64.24$\pm$0.43 & 37.82$\pm$0.39 & 47.23$\pm$0.44 \\
    
    COINCIDE & 
    22.56$\pm$0.90 & 29.89$\pm$1.42 & 44.56$\pm$0.84 & 35.83$\pm$0.48 & 
    69.02$\pm$0.59 & 66.82$\pm$0.69 & 39.14$\pm$0.85 & 49.42$\pm$0.82 \\
    
    DBP & 
    20.28$\pm$0.81 & 28.67$\pm$0.32 & 43.25$\pm$0.45 & 34.75$\pm$0.54 &
    66.74$\pm$0.82 & 64.95$\pm$0.06 & 35.93$\pm$0.26 & 45.86$\pm$0.98 \\

    InfoBatch & 
    22.93$\pm$0.73 & 29.14$\pm$0.56 & 47.65$\pm$0.30 & 40.60$\pm$0.96 &
    \underline{74.38$\pm$0.37} & 69.84$\pm$0.24 & 41.26$\pm$0.27 & 50.19$\pm$0.13 \\

    DivBS & 
    23.41$\pm$0.14 & 31.72$\pm$0.18 & 47.25$\pm$0.74 & 40.38$\pm$0.62 &
    \underline{74.37$\pm$0.75} & \underline{70.65$\pm$0.57} & 44.25$\pm$0.08 & \underline{52.23$\pm$0.76} \\

    TIVE & 
    21.15$\pm$0.09 & 28.12$\pm$0.10 & 44.72$\pm$0.58 & 35.28$\pm$1.37 &
    68.15$\pm$0.41 & 65.79$\pm$0.90 & 36.57$\pm$0.84 & 47.04$\pm$0.02 \\

    Adapt-$\infty$ &
    22.42$\pm$0.16 & 29.39$\pm$0.72 & 45.06$\pm$0.34 & 35.51$\pm$1.29 & 
    71.48$\pm$0.97 & 68.45$\pm$0.31 &
    42.38$\pm$0.14 & 50.48$\pm$0.44 \\

    \cmidrule(lr){1-1} \cmidrule(lr){2-3} \cmidrule(lr){4-5} \cmidrule(lr){6-7} \cmidrule(lr){8-9} 
    
    \ours (\textbf{Ours}) & 
    \textbf{25.67}$\pm$\textbf{0.35} & \textbf{34.23}$\pm$\textbf{0.38} & \textbf{49.98}$\pm$\textbf{0.27} & \textbf{43.94}$\pm$\textbf{0.31} & \textbf{75.26}$\pm$\textbf{0.35} & \textbf{71.91}$\pm$\textbf{0.55} &
    \textbf{45.56}$\pm$\textbf{0.35} & \textbf{53.48}$\pm$\textbf{0.17} \\

    \bottomrule
    \end{tabular}
    }
    \vspace{-0.5em}
    \caption{
    \textbf{Quantitative comparison across MLLM and LLM benchmarks at a 6.25\% selection ratio.}
    We employ LLaVA-1.5-7B (MLLM) and LLaMA-3.1-8B (LLM). 
    Bold indicates the highest performance; underlined results are within the 0.05 t-test significance level.
    'Full-Data Training' uses all data without selection.
    % \ours outperforms the baselines across various selection ratios.
    % Bold indicates the highest performance, and underlined results are within the 0.05 significance level of the t-test.
    % `Full Data Training' denotes training the MLLM on all incoming data without data selection.
    % LLaVA-1.5-7B is used as the MLLM.
    % LLaVA-1.5-7B is used.
    % Bold indicates the highest performance; underlined results are within the 0.05 t-test significance level. 
    }
  \label{tab:main_llava_coast_adapt_625}
  \vspace{-.5em}
\end{table*}

\noindent \textbf{Comparison of Computation Budget.}
\begin{figure}[ht!]
    \vspace{-0.7em}
    \centering
    \includegraphics[width=0.95\linewidth]{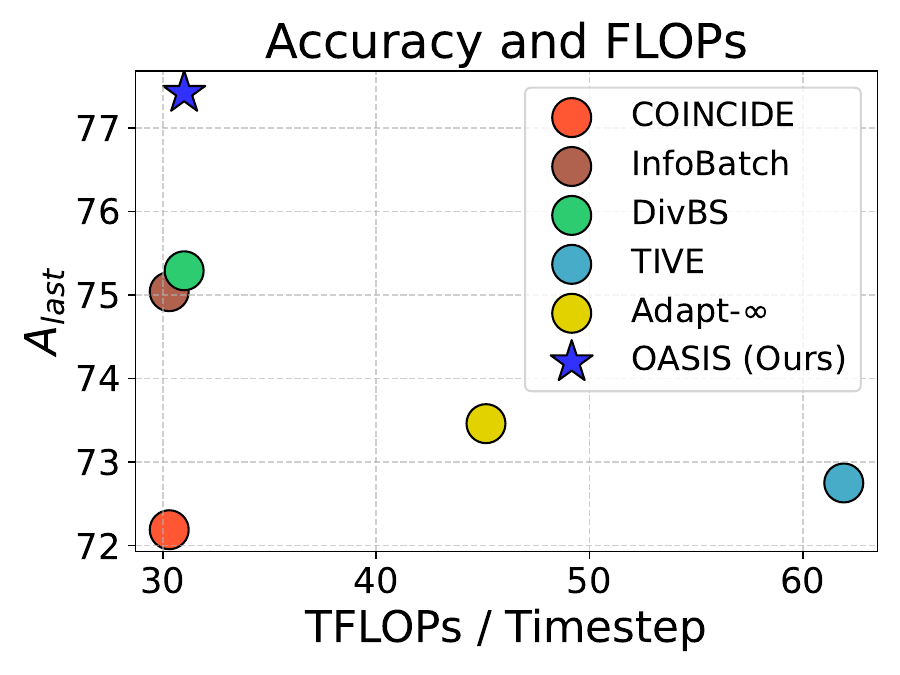}
    \vspace{-1.0em}
    \caption{
    \textbf{Accuracy and FLOPs with 25\% selection ratio on \oursbenchmark.} 
    The top-left corner illustrates effective and efficient sample selection.
    }
    \label{fig:comp_flops}
    \vspace{-1em}
\end{figure}
We compare the computational cost of \ours and baselines, as shown in Fig.~\ref{fig:comp_flops}.
\ours achieves higher performance while requiring fewer FLOPs.
See Sec.~\ref{sec:appx:cost_comparison} for a detailed computational analysis.

\noindent \textbf{Comparison of Selected Samples' Diversity.}
\begin{table}[ht!]
    \centering
    \resizebox{0.85\linewidth}{!}{
    \begin{tabular}{lcc}
    \toprule
    \multirow{1}{*}{Method} & 
    \multicolumn{1}{c}{\oursbenchmark} & \multicolumn{1}{c}{COAST} \\ 
    % \cmidrule(lr){2-2} \cmidrule(lr){3-3}
    \cmidrule(lr){1-1} \cmidrule(lr){2-2} \cmidrule(lr){3-3}
    Random & 0.263 & 0.624 \\
    GradNorm {\footnotesize \color{blue}{(ICML 2018)}} & 0.255 & 0.598 \\
    Self-Sup {\footnotesize \color{blue}{(NeurIPS 2022)}} & 0.358 & 0.581 \\
    COINCIDE {\footnotesize \color{blue}{(EMNLP 2024)}} & 0.259 & 0.604 \\
    DBP {\footnotesize \color{blue}{(ICLR 2024)}} & 0.288 & 0.587 \\
    InfoBatch {\footnotesize \color{blue}{(ICLR 2024)}} & 0.308 & 0.632 \\
    DivBS {\footnotesize \color{blue}{(ICML 2024)}} & 0.291 & 0.625 \\
    TIVE {\footnotesize \color{blue}{(arXiv:2403)}} & 0.296 & 0.613 \\
    Adapt-$\infty$ {\footnotesize \color{blue}{(ICLR 2025)}} & 0.274 & 0.578 \\

    \cmidrule(lr){1-1} \cmidrule(lr){2-2} \cmidrule(lr){3-3}
    \ours w/o \ourstwo & 0.261 & 0.599 \\
    \ours (Ours)     & \textbf{0.242} & \textbf{0.563} \\

    \bottomrule
    \end{tabular}
    }
    \vspace{-0.3em}
    \captionof{table}{
    \textbf{Density comparison of selected samples.}
    % We measure densities of all selected samples by each selection method. 
    Lower density indicates higher diversity.
    }
    \vspace{-1em}
    \label{tab:diversity}
\end{table}
We quantify diversity using kernel density, computed as the mean pairwise similarity under a Gaussian kernel, following \citet{lee2024concept}.
% ; lower density indicates higher diversity. 
As shown in Tab.~\ref{tab:diversity}, \ours selects the most diverse set among all methods.
We attribute this to the probabilistic sampling in \oursone, which promotes diversity, and the redundancy reduction in \ourstwo.

% \begin{table}[h!]
%     \centering
%     \resizebox{0.85\linewidth}{!}{
%     \begin{tabular}{lcc}
%     \toprule
%     \multirow{1}{*}{Method} & 
%     \multicolumn{1}{c}{\oursbenchmark} & \multicolumn{1}{c}{COAST} \\ 
%     % \cmidrule(lr){2-2} \cmidrule(lr){3-3}
%     \cmidrule(lr){1-1} \cmidrule(lr){2-2} \cmidrule(lr){3-3}
%     Random & 0.263 & 0.624 \\
%     GradNorm {\footnotesize \color{blue}{(ICML 2018)}} & 0.255 & 0.598 \\
%     Self-Sup {\footnotesize \color{blue}{(NeurIPS 2022)}} & 0.358 & 0.581 \\
%     COINCIDE {\footnotesize \color{blue}{(EMNLP 2024)}} & 0.259 & 0.604 \\
%     DBP {\footnotesize \color{blue}{(ICLR 2024)}} & 0.288 & 0.587 \\
%     InfoBatch {\footnotesize \color{blue}{(ICLR 2024)}} & 0.308 & 0.632 \\
%     DivBS {\footnotesize \color{blue}{(ICML 2024)}} & 0.291 & 0.625 \\
%     Adapt-$\infty$ {\footnotesize \color{blue}{(ICLR 2025)}} & 0.274 & 0.578 \\

%     \cmidrule(lr){1-1} \cmidrule(lr){2-2} \cmidrule(lr){3-3}
%     \ours (Ours)     & \textbf{0.242} & \textbf{0.563} \\

%     \bottomrule
%     \end{tabular}
%     }
%     \caption{
%     \textbf{Density comparison of selected samples.}
%     % We measure densities of all selected samples by each selection method. 
%     Lower density indicates higher diversity.
%     }
%     \label{tab:diversity}
% \end{table}

\begin{figure*}
    \vspace{-.75em}
    \centering   
    \includegraphics[width=\linewidth]{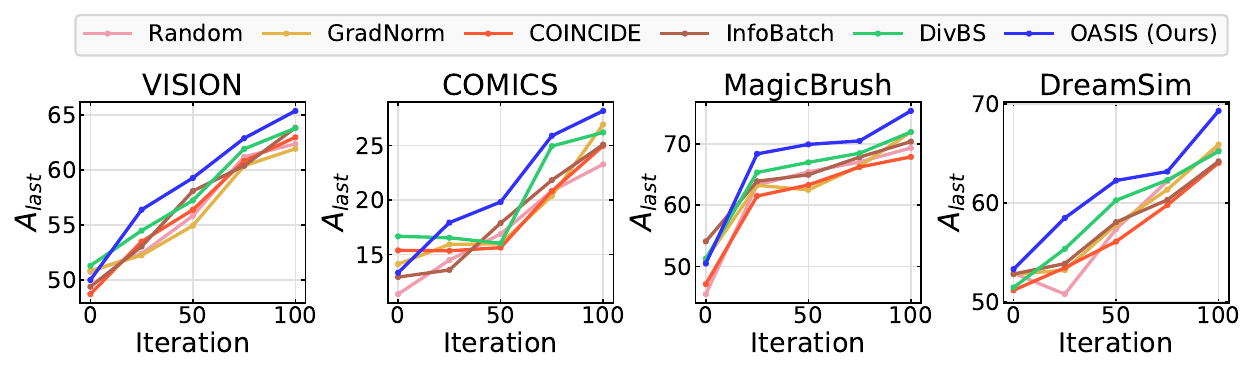}
    
    % {\vspace{-0.5em} \footnotesize (a) \oursbenchmark}
    
    % \centering   
    % \includegraphics[width=\linewidth]{figures/fewshot_coast.pdf}
    
    % {\vspace{-0.5em} \footnotesize (b) COAST}
    \vspace{-0.85em}
    \caption{
    \textbf{Comparison of fast adaptation performance.}
    After CIT of LLaVA-1.5-7B on subsets (25\% of the full data), selected using each sample selection baseline from \oursbenchmark, we fine-tune the model for 100 epochs on each downstream task (\ie, VISION, COMICS, MagicBrush, and DreamSim). 
    }
    \label{fig:fewshot}
\end{figure*}

\begin{table*}[t]
  % \label{tab:headings}
    \vspace{-0.3em}
  \centering
  \resizebox{\linewidth}{!}{
    \begin{tabular}{lcccccc}
    \toprule
    \multirow{2}{*}{Method} & \multicolumn{2}{c}{\oursbenchmark} & \multicolumn{2}{c}{COAST} & \multicolumn{2}{c}{Adapt} \\
    
    & $A_{avg} \ \uparrow$ & $A_{last} \ \uparrow$ & $A_{avg} \ \uparrow$ & $A_{last} \ \uparrow$ & $A_{avg} \ \uparrow$ & $A_{last} \ \uparrow$ \\ 

    \cmidrule(lr){1-1} \cmidrule(lr){2-3} \cmidrule(lr){4-5} \cmidrule(lr){6-7} 

    % Increased Iterations
    Vanilla & 
    62.54$\pm$0.55 & 67.16$\pm$0.39 &
    21.35$\pm$0.38 & 29.39$\pm$0.26 &
    46.36$\pm$0.53 & 38.57$\pm$0.38 \\

    (+) \oursone & 
    65.55$\pm$0.41 & 72.45$\pm$0.30 &
    25.80$\pm$0.32 & 35.28$\pm$0.52 &
    50.52$\pm$0.42 & 44.65$\pm$0.10 \\

    (+) \oursone \& \ourstwo (\textbf{Ours}) & 
    \textbf{67.58}$\pm$\textbf{0.46} & \textbf{74.51}$\pm$\textbf{0.48} &
    \textbf{27.83}$\pm$\textbf{0.50} & \textbf{37.29}$\pm$\textbf{0.13} &
    \textbf{51.33}$\pm$\textbf{1.16} & \textbf{46.22}$\pm$\textbf{0.57} \\
    \bottomrule
    \end{tabular}
    }
    \vspace{-0.3em}
    \caption{
    \textbf{Ablations for proposed components of \ours.}
    We train LLaVA-1.5-7B using 6.25\% of the dataset samples with 0.125, 0.25, and 0.25 online iterations in \oursbenchmark, COAST, and Adapt, respectively.
    Vanilla: selecting a fixed number of samples with the highest FI per batch.
    }
  \label{tab:main_ablation}
  \vspace{-.6em}
\end{table*}

\noindent \textbf{Comparison of Fast Adaptation Performance.}
We evaluate fast adaptation on unseen downstream tasks following CIT, where CIT serves as upstream continual pre-training.
This setup assesses the generalizability of models trained on subsets selected by different online sample selection methods.
Downstream tasks include COMICS~\citep{iyyer2017amazing}, VISION~\citep{bai2023vision}, DreamSim~\citep{fu2023dreamsim}, and MagicBrush~\citep{Zhang2023magicbrush}, while \oursbenchmark serve as upstream datasets.
Results in Fig.~\ref{fig:fewshot} show that subsets selected by \ours yield consistently superior performance across all downstream tasks, demonstrating superior generalizability.
Note that zero-shot performance (\ie, iteration 0) can vary depending on the similarity between the downstream tasks and the subset of data selected by each baseline, resulting in inconsistent ordering between baselines.
However, as fine-tuning on downstream tasks progresses, \ours exhibits consistently faster adaptation, demonstrating the superior generalizability of models trained on subsets selected by \ours.
See Sec.~\ref{sec:appx:appen_few-shot} for additional fast adaptation results across benchmarks.

% \noindent \textbf{Detailed Studies.}
% % Comparison of Information Metrics
% Moreover, we provide detailed studies in the appendix for space's sake. 
% Specifically, we provide a comparison of informativeness metrics in Sec.~\ref{sec:appx:info_exp}, 

\subsection{Ablation Study}

% \paragraph{Ablation Study on Different Components.}
\noindent \textbf{Ablation Study on Different Components.}
We ablate \ours to investigate the benefit of each proposed component, and summarize the results in Tab.~\ref{tab:main_ablation}.
As shown, \oursone significantly improves performance by identifying informative samples, while \ourstwo further enhances performance by reducing redundancy between selected samples.
Consistent improvements across benchmarks validate the effectiveness of our proposed components.

% We provide 

% \noindent \textbf{Ablation Study on Task Order.}
% We also present an ablation study on the effects of varying task orders across two benchmarks, COAST and MICVIT, in Sec.~\ref{sec:appx:ablation_taskorder}, where we observe that different task orders have minimal impact on overall performance.

\noindent \textbf{Additional Analyses.}
% Comparison of Information Metrics
% Moreover, we include additional analyses in the appendix. 
We further evaluate \ours across different model scales (\eg, 0.5B, 1B, 3B, and 13B) (Sec.~\ref{sec:appx:scale_exp}) and analyze informativeness metric evaluations (Sec.~\ref{sec:appx:info_exp}), effect of task orders (Sec.~\ref{sec:appx:ablation_taskorder}), effect of EMA ratio $\beta$ (Sec.~\ref{sec:appx:ema_ratio}), accuracy over time (Sec.~\ref{sec:appx:avg_accuracy}), and comparison of the number of selected samples (Sec.~\ref{sec:appx:num_samples}).

% Specifically, we investigate the effects of varying task orders (Sec.~\ref{sec:appx:ablation_taskorder}), the effect of the EMA ratio $\alpha$ (Sec.~\ref{sec:appx:ema_ratio}), comparison of computational cost between \ours and baselines (Sec.~\ref{sec:appx:cost_comparison}), comparison of informativeness metrics (Sec.~\ref{sec:appx:info_exp}), and comparison of accuracy over time in CVIT (Sec.~\ref{}).
% highlighting the benefits of using Fisher Information in our framework.

% ???
% Additional experimental results on the benchmarks are also included in the appendix.

\section{Conclusion}
We address the challenge of achieving high performance while training on a small subset of data in the online CIT setup. 
Prior works select a fixed number of samples per batch, 
which struggles in the CIT setup due to varying informativeness and frequent redundancy from co-occurring similar samples.
To address this, we propose \ours, comprising two components: \oursone, which selects informative samples by estimating each sample’s informativeness relative to all previously encountered data, and \ourstwo, which reduces redundancy via sample-wise similarity.
Extensive experiments on diverse CIT benchmarks demonstrate that our proposed \ours consistently selects informative samples, outperforms existing baselines, and achieves performance comparable to full-data training.

% We tackle the challenge of achieving high performance while training on a small subset of data in the online CVIT setting. Prior methods select a fixed number of samples per batch, which fails under CVIT due to varying informativeness and frequent redundancy from co-occurring similar samples. To address this, we propose \ours, consisting of two components: \oursonefull, which adaptively selects samples based on relative informativeness, and \ourstwofull, which mitigates redundancy via sample-wise similarity. Extensive experiments on diverse CVIT benchmarks show that \ours consistently selects informative samples, outperforms existing baselines, and approaches full-data performance.

%  online sample selection

\noindent \paragraph{Limitations.}
Our method efficiently selects informative samples in CIT, requiring only a forward pass without any backward computation.
A promising future direction is to remove this forward-pass requirement entirely, enabling near-instantaneous adaptation and substantially faster processing for real-time CIT applications.

% \section*{References}

\bibliography{custom}

\appendix
\clearpage

% \section{Example Appendix}
% \label{sec:appendix}

% This is an appendix.
\section{Technical Appendices and Supplementary Material}
\subsection{Proof of Theorem~\ref{thm:main_normal}}
\label{sec:appx:proof_thm1}
In continual instruction tuning (CIT), the data distribution evolves gradually over time (\eg, more Christmas-related samples in winter and more air-conditioner-related samples in summer)~\citep{koh2023online}. 
Although this constitutes a distribution shift, within short horizons (\eg, a week), the distribution of training batches can be regarded as approximately stable under memory-only training~\citep{koh2021online, seo2024just}, even in the presence of sharp task boundaries.
Specifically, memory-only training constructs batches by only retrieving from an episodic memory containing previously encountered samples, thereby inducing smoother changes of training distribution, unlike ER~\citep{rolnick2019experience}, which mixes continuously encountered streaming data with memory samples and may cause abrupt shifts.
As a result, our choice of memory-only training for CIT motivates the local stationarity assumption (Assumption~\ref{asmp:local_stationarity}) and local weak dependence assumption (Assumption~\ref{asmp:local_weakdep}) introduced below.

\paragraph{Setting.}
At time $t$, we observe a batch $\{(x^{(t)}_i, y^{(t)}_i)\}_{i=1}^{N_\mathcal{B}}$ and calculate sample-wise FI $\{I^{(t)}_i\}_{i=1}^{N_\mathcal{B}}$ using Fisher Information Matrix (Eq.~\ref{eq:select}), while updating EMA $\mu_t$ and EMV $\sigma_t$ of FI (Eq.~\ref{eq:EMA_EMV}) using FI averaged by batch $(\bar{I}^{(t)}=\frac{1}{N_\mathcal{B}}\sum_{i=1}^{N_\mathcal{B}}I^{(t)}_i)$.
We decompose $\bar I^{(t)}$ as
\[
\bar I^{(t)} = m_t + \varepsilon_t,
\]
where $m_t:=\E[\bar I^{(t)}]$ denotes the underlying (population) FI at time $t$, and 
$\varepsilon_t:=\bar I^{(t)} - m_t$ represents the centered sampling noise.

\begin{assumption}[Local Stationarity]
\label{asmp:local_stationarity}
Although the data distribution $P_t$ and model parameters $\theta_t$ evolve over time in continual learning, 
we assume that such changes are \emph{smooth}. 
In particular, within a short horizon comparable to the EMA effective window 
$L_{\mathrm{eff}} \approx \tfrac{1+\beta}{1-\beta}$, 
the mean $m_t=\E[\bar{I}^{(t)}]$ and variance $s_t^2=\Var(\bar{I}^{(t)})$ remain nearly constant. 
Formally, for temporal lag $h \in \mathbb{Z}$ with $|h|\le L_{\mathrm{eff}}$, 
\begin{gather}
\sup_{|h|\le L_{\mathrm{eff}}} |m_{t+h}-m_t| = o(\sqrt{1-\beta}), \\
\sup_{|h|\le L_{\mathrm{eff}}} |s_{t+h}^2-s_t^2| = o(1).
\end{gather}

% This reflects the intuition that distributional shifts (\eg, seasonal effects in real-world data) occur gradually rather than abruptly.
\end{assumption}

\begin{assumption}[Local Weak Dependence]
\label{asmp:local_weakdep}
The noise $\{\varepsilon_t\}$ is assumed to be \emph{locally weakly dependent}. 
That is, while sequential batches are correlated due to temporal proximity, 
correlations decay sufficiently fast as the temporal lag $h$ grows. Formally, for lag $h \to \infty$, 
\[
\Cov(\varepsilon_t, \varepsilon_{t+h}) \to 0.
\]
\end{assumption}

% ================== Chi-square structure: moments & rates ==================
\begin{lemma}[$\hat{I}^{(t)}$ has finite $(2+\delta)$-moments]
\label{lem:chi2_moments}
We estimate the Fisher information (FI) as the trace of the Fisher information matrix (FIM) in Eq.~\ref{eq:select}. 
Since the diagonal entries of the FIM are typically large in magnitude, 
the FI of a sample $(x^{(t)}_i, y^{(t)}_i) \in \mathcal{B}_t$, denoted $I^{(t)}_i$, can be approximated by a chi-square distribution~\citep{dauncey2024approximations}, 
\[
    I^{(t)}_i \;\overset{d}{\approx}\; \sigma_t^2\chi^2_{k_t},
\]
where $\sigma_t^2$ is a scale parameter chosen such that the mean and variance match those of the 
Fisher information trace, \ie,
\[
\mathbb{E}[I_i^{(t)}] \;\approx\; \sigma_t^2 k_t,
\quad 
\Var(I_i^{(t)}) \;\approx\; 2\sigma_t^4 k_t.
\]

Because both the sum of chi-square variables remains chi-square distributed, 
the batch-averaged FI $\bar{I}^{(t)}$ also follow chi-square approximations: 
\[
    \bar{I}^{(t)} \;\overset{d}{\approx}\; \frac{\sigma_t^2}{B}\chi^2_{k_t\mathcal{B}_t}.
\]

Since chi-square distributions have finite moments of all orders, it follows that for every $\delta > 0$,
\[
\E\!\left[\,|\bar{I}^{(t)}|^{2+\delta}\,\right] \;<\; \infty,
\quad
\E\!\left[\,|I^{(t)}_i|^{2+\delta}\,\right] \;<\; \infty.
\]
\end{lemma}

\begin{lemma}[Normal approximation rate for FI trace via chi-square]
\label{lem:chi2_rate}
Let $X\sim \chi^2_{k}$ and consider the standardized variable
$\widetilde X=(X-k)/\sqrt{2k}$. Then classical Berry-Esseen-type results (~\citet{dasgupta2008asymptotic, chen2010normal}) guarantee that
\[
\sup_{x\in\mathbb R}\big|\Pr(\widetilde X\le x)-\Phi(x)\big|
\;=\; O\!\left(\frac{1}{\sqrt{k}}\right).
\]
Applied to $\bar I^{(t)}$ with $\bar{I}^{(t)} \approx \frac{\sigma_t^2}{\mathcal{B}_t}\chi^2_{k_t\mathcal{B}_t}$, we obtain
\begin{align}
\sup_{x}\Bigg|\Pr\!\Bigg(
\frac{\bar I^{(t)} - \E[\bar I^{(t)}]}
     {\sqrt{\Var(\bar I^{(t)})}} \le x
\Bigg) - \Phi(x)\Bigg| \nonumber \\
= O\!\left(\frac{1}{\sqrt{k_t \mathcal{B}_t}}\right).
\end{align}
% If one instead standardizes the single-sample FI $I^{(t)}_i$ under a $\chi^2_{k_t}$-type model, the same $O(k_t^{-1/2})$ rate holds.
\end{lemma}

\begin{lemma}[EMA/EMV are consistent approximations of the true mean and variance]
\label{lem:ratio_consistency_easy}
Let $m_t = \E[\bar I^{(t)}]$ be the true mean of the batch-averaged FI and 
$s_t^2 = \Var(I^{(t)}_i)$ the true sample-wise variance at time $t$. 
Let $\mu_{t-1}$ and $v_{t-1}$ be the EMA and EMV computed from past batches, and define 
$\widehat\sigma_{t-1}^2 := B\cdot v_{t-1}$.
Then, under Assumptions~\ref{asmp:local_stationarity} and \ref{asmp:local_weakdep},
\[
|\mu_{t-1}-m_t| = o(s_t), 
\qquad 
\frac{\widehat\sigma_{t-1}^2}{s_t^2} \;\xrightarrow{p}\; 1.
\]

\textbf{In words:} 
the EMA mean $\mu_{t-1}$ is essentially the same as the true mean $m_t$, 
and the EMV-based variance estimator $\widehat\sigma_{t-1}^2$ consistently estimates the true variance $s_t^2$.
\end{lemma}

\begin{proof}[Proof Lemma~\ref{lem:ratio_consistency_easy}.]
\textbf{Mean part.}  
The EMA is a weighted moving average:
\[
\mu_{t-1} = (1-\beta)\sum_{k\ge1}\beta^{k-1}\bar I_{t-k}.
\]
Local stationarity (Assumption~\ref{asmp:local_stationarity}) says that the true mean $m_t$ does not change much 
over the EMA window $L_{\mathrm{eff}}\approx \tfrac{1+\beta}{1-\beta}$. 
Hence each difference $|m_{t-k}-m_t|$ is small, and the exponential weights $(1-\beta)\beta^{k-1}$ quickly decay. 
As a result, the EMA bias
\[
|\mu_{t-1}-m_t| = (1-\beta)\sum_{k\ge1}\beta^{k-1}|m_{t-k}-m_t|
\]
is negligible compared to the natural variability $s_t$; formally, it is $o(s_t)$.

\textbf{Variance part.}  
The EMV is defined as
\[
v_{t-1} = (1-\lambda)\sum_{k\ge1}\lambda^{k-1}(\bar I_{t-k}-\mu_{t-k})^2.
\]
This is an exponentially weighted average of past squared deviations.  
Local weak dependence (Assumption~\ref{asmp:local_weakdep}) ensures that correlations between far-apart batches vanish, 
so $v_{t-1}$ consistently estimates $\Var(\bar I_t)$ (this is the usual consistency of exponential-kernel HAC estimators).
Finally, since $\Var(\bar I_t)=s_t^2/B$ under i.i.d.\ samples in the batch, 
multiplying $v_{t-1}$ by $B$ yields a consistent estimator of the sample-wise variance $s_t^2$.

Thus, both the EMA mean and the EMV variance converge to their true population counterparts.
\end{proof}

\begin{proof}[Proof of Theorem~\ref{thm:main_normal}]

We prove that the one-step-lag standardized statistic
\(
Z_{t,i}=\frac{I^{(t)}_i-\mu_{t-1}}{\widehat\sigma_{t-1}}
\)
is well-approximated by $\mathcal N(0,1)$ under 
Assumptions~\ref{asmp:local_stationarity}--\ref{asmp:local_weakdep} and the chi-square structure of FI traces.

\paragraph{Step 1: Within-batch normal approximation.}
By Lemma~\ref{lem:chi2_moments}, the sample-wise FI admits a (scaled) chi-square representation:
\(I^{(t)}_i \overset{d}{\approx} \sigma_t^2\,\chi^2_{k_t}\)
for some effective degrees-of-freedom $k_t$.
Therefore, by Lemma~\ref{lem:chi2_rate},
\begin{equation}
\label{eq:within_normal}
\sup_{x\in\mathbb R}
\Big|
\Pr\!\Big(\tfrac{I^{(t)}_i - m_t}{s_t}\le x\Big) - \Phi(x)
\Big|
= O\!\big(k_t^{-1/2}\big),
\end{equation}
where $m_t=\E[\bar I_t]=\E[I^{(t)}_i]$ and $s_t^2=\Var(I^{(t)}_i)$.

\paragraph{Step 2: EMA/EMV ratio-consistency.}
The ideal normalization requires $(m_t,s_t)$, but in practice we use 
the EMA mean $\mu_{t-1}$ and EMV variance $\widehat\sigma_{t-1}^2$ from past batches. 
By Lemma~\ref{lem:ratio_consistency_easy}, local stationarity ensures that 
$|\mu_{t-1}-m_t|=o(s_t)$, and local weak dependence guarantees 
$\widehat\sigma_{t-1}^2/s_t^2 \xrightarrow{p} 1$. 
Hence $(\mu_{t-1},\widehat\sigma_{t-1})$ consistently approximate $(m_t,s_t)$, 
justifying their use as practical normalizers.

% \paragraph{Step 2: EMA/EMV ratio-consistency (replacement).}
% We now replace the (unknown) $(m_t,s_t)$ by the operational normalizers $(\mu_{t-1},\widehat\sigma_{t-1})$ built from \emph{past} batches, following Lemma~\ref{lem:ratio_consistency_easy}.
% % We use the following standard fact (proved just below):

\paragraph{Step 3: Slutsky/self-normalization.}
We can decompose $Z_{t,i}$ as
\begin{equation}
\label{eq:decomp}
Z_{t,i}
= \frac{I^{(t)}_i - m_t}{s_t}\cdot \frac{s_t}{\widehat\sigma_{t-1}}
\;-\; \frac{\mu_{t-1}-m_t}{\widehat\sigma_{t-1}}.
\end{equation}
By \eqref{eq:within_normal}, the standardized term $(I^{(t)}_i-m_t)/s_t$ 
is $O(k_t^{-1/2})$, close to $\mathcal N(0,1)$ in Kolmogorov distance. 
Lemma~\ref{lem:ratio_consistency_easy} further ensures that 
$s_t/\widehat\sigma_{t-1}\xrightarrow{p}1$ and 
$(\mu_{t-1}-m_t)/\widehat\sigma_{t-1}=o_p(1)$. 
Hence, both adjustment factors vanish asymptotically, and by Slutsky’s theorem 
the distribution of $Z_{t,i}$ retains the normal approximation:
\[
\sup_{x\in\mathbb R}\Big|\Pr(Z_{t,i}\le x)-\Phi(x)\Big|
\;\le\; C_1\,k_t^{-1/2} \;+\; o(1).
\]

\paragraph{Step 4: One-step lag and leakage.}
We used $(\mu_{t-1},\widehat\sigma_{t-1})$ built from \emph{past} batches only, 
so $I^{(t)}_i$ is not used inside its own normalizer; hence no self-normalization bias occurs.
(If one includes $t$ in the EMA/EMV, the induced bias is $O(1-\beta)$.)

Combining the steps establishes the stated claim \(Z_{t,i}\approx\mathcal N(0,1)\) and the quantitative bound.
\end{proof}

% \begin{lemma}[EMA/EMV ratio-consistency on sample scale]
% \label{lem:ratio_consistency_proof}
% Under Assumptions~\ref{asmp:local_stationarity}--\ref{asmp:local_weakdep},
% \(
% |\mu_{t-1}-m_t| = o(s_t)
% \)
% and
% \(
% \widehat\sigma_{t-1}^2/s_t^2 \xrightarrow{p} 1
% \),
% where $\widehat\sigma_{t-1}^2 := B\cdot v_{t-1}$ in the i.i.d.\ within-batch case 
% (or $B_{\mathrm{eff}}\cdot v_{t-1}$ with within-batch correlation).
% \end{lemma}

% \noindent\emph{Proof of Lemma~\ref{lem:ratio_consistency_proof}.}
% Local stationarity (Assumption~\ref{asmp:local_stationarity}) implies that on the EMA window 
% of length $L_{\mathrm{eff}}\approx \frac{1+\beta}{1-\beta}$,
% the mean $m_t$ drifts smoothly.
% Therefore, the EMA bias satisfies
% \(
% |\mu_{t-1}-m_t|
% =(1-\beta)\sum_{k\ge1}\beta^{k-1}|m_{t-k}-m_t|
% =o(s_t),
% \)
% since the weighted differences vanish at a rate dominated by $s_t$.

% For the variance, let $w_k=(1-\lambda)\lambda^{k-1}$ be EMV weights and write the 
% past batch-average deviations as $(\bar I_{t-k}-\mu_{t-k})$.
% Local weak dependence (Assumption~\ref{asmp:local_weakdep}) along with local stationarity ensures that $v_{t-1}$ is a consistent estimator of $\Var(\bar I_t)$ on the current window (exponential-kernel HAC consistency).
% Since $\Var(\bar I_t)=s_t^2/B$ under within-batch i.i.d.\ (or $s_t^2/B_{\mathrm{eff}}$ with correlation),
% multiplying by $B$ (or $B_{\mathrm{eff}}$) yields $\widehat\sigma_{t-1}^2\stackrel{p}{\to} s_t^2$. \hfill$\square$

\subsection{Implementation Details}
\label{sec:appx:implementation_details}

We fine-tune models using learning rates from their original papers for one epoch: 2e-5 for both LLaVA-1.5-7B and Qwen-VL-2.5-7B and 3e-4 for both LLaMA-3.1-8B and Qwen3-8B.
We use the Adam optimizer with no decay and a Cosine LR scheduler. 
For batch size $N_\mathcal{B}$, we use 16 for all experiments.
For LoRA finetuning, we adopt LoRA with rank 128 at all linear layers in the LLM backbone.
We set the number of batch iterations per sample encounter to COAST, Adapt, and \oursbenchmark as 0.125, 0.125, and 0.0625, respectively.
We assume an infinite memory setup, which assumes that all encountered samples can be stored in an episodic memory.
For the batch retrieval method, we adopt memory-only training~\citep{koh2021online, seo2024learning}, where training batches are only retrieved from the episodic memory at each iteration, enhancing robustness to distribution shifts~\citep{koh2023online, seo2025budgeted}.
We set the EMA ratio $\beta$ to 0.9 for all datasets.
We conduct experiments on NVIDIA RTX A6000 GPUs. Every experiment takes no more than 2 days.

\subsection{Experimental Results on Various Benchmarks}
\label{sec:appx:llava_benchmark_results}

In addition to \oursbenchmark, we compare \ours with sample selection methods in COAST~\citep{cao2024continual} and Adapt~\citep{maharana2025adaptinfty} under selection ratios of 12.5\% and 25.0\% in Tab.~\ref{tab:main_llava_coast_adapt_125_25}.
\ours not only outperforms selection baselines but also achieves comparable results with a model trained on all data (\ie, Full-Data Training) in diverse benchmarks.

\begin{table*}[t]
  \centering
  \resizebox{\linewidth}{!}{
    \begin{tabular}{lcccccccc}
    \toprule
    \multirow{3}{*}{Method} 
    & \multicolumn{8}{c}{Selection Ratio (\%)} \\ 
    \cmidrule(lr){2-9}
    & \multicolumn{4}{c}{12.5} & \multicolumn{4}{c}{25.0} \\ 
    \cmidrule(lr){2-5} \cmidrule(lr){6-9}
    & \multicolumn{2}{c}{COAST} & \multicolumn{2}{c}{Adapt} 
    & \multicolumn{2}{c}{COAST} & \multicolumn{2}{c}{Adapt} \\ 
    & $A_{avg} \ \uparrow$ & $A_{last} \ \uparrow$ & $A_{avg} \ \uparrow$ & $A_{last} \ \uparrow$ 
    & $A_{avg} \ \uparrow$ & $A_{last} \ \uparrow$ & $A_{avg} \ \uparrow$ & $A_{last} \ \uparrow$ \\ 
    \cmidrule(lr){1-1} \cmidrule(lr){2-3} \cmidrule(lr){4-5} \cmidrule(lr){6-7} \cmidrule(lr){8-9}

    Full-Data Training & 
    31.56$\pm$1.42 & 39.06$\pm$0.55 &
    55.89$\pm$0.20 & 54.26$\pm$0.23 &
    31.56$\pm$1.42 & 39.06$\pm$0.55 &
    55.89$\pm$0.20 & 54.26$\pm$0.23 \\

    \cmidrule(lr){1-1} \cmidrule(lr){2-3} \cmidrule(lr){4-5} \cmidrule(lr){6-7} \cmidrule(lr){8-9}

    Random & 
    24.58$\pm$0.17 & 32.45$\pm$0.92 &
    49.04$\pm$0.49 & 44.19$\pm$0.62 &
    25.82$\pm$0.10 & 34.28$\pm$0.28 &
    50.68$\pm$0.30 & 47.58$\pm$0.34 \\

    GradNorm {\footnotesize \color{blue}{(ICML 2018)}} & 
    23.35$\pm$0.38 & 29.44$\pm$0.28 &
    48.47$\pm$0.63 & 45.36$\pm$0.62 &
    25.67$\pm$0.21 & 32.85$\pm$0.36 &
    51.29$\pm$0.54 & 48.96$\pm$0.81 \\

    Self-Sup {\footnotesize \color{blue}{(NeurIPS 2022)}} & 
    23.72$\pm$0.27 & 31.37$\pm$0.75 &
    47.04$\pm$0.26 & 43.94$\pm$1.38 &
    25.17$\pm$0.37 & 32.84$\pm$0.42 &
    49.34$\pm$0.85 & 46.46$\pm$0.47 \\

    COINCIDE {\footnotesize \color{blue}{(EMNLP 2024)}} & 
    24.45$\pm$0.84 & 31.08$\pm$0.59 &
    48.52$\pm$0.55 & 44.76$\pm$0.43 &
    26.22$\pm$0.34 & 34.90$\pm$0.57 &
    51.74$\pm$0.64 & 47.40$\pm$0.31 \\

    DBP {\footnotesize \color{blue}{(ICLR 2024)}} & 
    22.87$\pm$0.33 & 25.81$\pm$0.37 &
    46.24$\pm$0.36 & 43.30$\pm$1.08 &
    24.74$\pm$0.70 & 32.13$\pm$0.64 &
    48.58$\pm$0.43 & 45.65$\pm$0.84 \\

    InfoBatch {\footnotesize \color{blue}{(ICLR 2024)}} & 
    24.78$\pm$0.58 & 31.74$\pm$0.62 &
    \underline{49.77$\pm$0.82} & \underline{47.54$\pm$0.36} &
    25.30$\pm$0.35 & 34.49$\pm$0.45 &
    51.55$\pm$0.47 & 49.04$\pm$0.77 \\

    DivBS {\footnotesize \color{blue}{(ICML 2024)}} & 
    25.13$\pm$0.12 & 33.30$\pm$0.87 &
    49.96$\pm$0.48 & 47.17$\pm$0.51 &
    26.91$\pm$0.02 & 35.01$\pm$0.16 &
    52.36$\pm$0.05 & \underline{50.02$\pm$1.21} \\

    TIVE {\footnotesize \color{blue}{(arXiv:2403)}} & 
    23.41$\pm$0.23 & 30.17$\pm$0.43 &
    47.75$\pm$0.73 & 44.38$\pm$0.99 &
    25.18$\pm$0.78 & 34.08$\pm$0.60 &
    49.69$\pm$1.24 & 45.97$\pm$0.88 \\

    Adapt-$\infty$ {\footnotesize \textcolor{blue}{(ICLR 2025)}} &
    24.73$\pm$0.36 & 31.69$\pm$0.35 &
    48.38$\pm$0.20 & 45.42$\pm$0.46 &
    26.07$\pm$0.25 & 33.50$\pm$0.04 &
    50.56$\pm$0.25 & 47.54$\pm$0.53 \\

    \cmidrule(lr){1-1} \cmidrule(lr){2-3} \cmidrule(lr){4-5} \cmidrule(lr){6-7} \cmidrule(lr){8-9}

    \ours (\textbf{Ours}) & 
    \textbf{27.13}$\pm$\textbf{0.70} & \textbf{35.42}$\pm$\textbf{0.49} &
    \textbf{51.73}$\pm$\textbf{0.33} & \textbf{49.02}$\pm$\textbf{0.85} &
    \textbf{28.72}$\pm$\textbf{0.63} & \textbf{37.55}$\pm$\textbf{0.28} &
    \textbf{54.58}$\pm$\textbf{0.14} & \textbf{51.87}$\pm$\textbf{0.49} \\

    \bottomrule
    \end{tabular}
  }
  \vspace{-0.3em}
  \caption{
  \textbf{Quantitative comparison on COAST and Adapt under different selection ratios (12.5\% and 25.0\%).}
  LLaVA-1.5-7B is used as the MLLM.
  Bold indicates the highest performance; underlined results are within the 0.05 t-test significance level.
  'Full-Data Training' uses all data without selection.
  }
  \label{tab:main_llava_coast_adapt_125_25}
  \vspace{-.6em}
\end{table*}

\subsection{Benchmark Configuration Details}
\label{sec:appx:benchmark_details}
% SuperNI benchmark comprises of total 15 tasks from the original 1,616 tasks by selecting 3 tasks from each of the 5 NLP task types: dialogue generation, information extraction, question answering, summarization, and sentiment analysis.
We use Long Sequence~\citep{razdaibiedina2023progressive} and TRACE~\citep{wang2023trace} as our text-only benchmarks.
Long Sequence consists of datasets from two existing CL benchmarks, GLUE~\citep{wang2018glue} and SuperGLUE~\citep{wang2019superglue}, along with additional IMDB movie reviews dataset~\citep{maas2011learning}.
TRACE benchmark consists of 8 tasks, NumGLUE-ds~\citep{mishra2022numglue}, ScienceQA~\citep{lu2022learn}, Py150~\citep{lu2021codexglue}, C-STANCE~\citep{zhao2023c}, FOMC~\citep{shah2023trillion}, 20Minuten~\citep{gonzales2021new}, NumGLUE-cm~\citep{mishra2022numglue}, and MeetingBank~\citep{hu2023meetingbank}, encompassing domain-specific tasks, multilingual capabilities, code generation, and mathematical reasoning.

We use COAST~\citep{cao2024continual}, Adapt~\citep{maharana2025adaptinfty}, and \oursbenchmark for our multi-modal benchmarks.
For the COAST benchmark, we use the COAST-domain, which emulates a scenario where MLLMs needs to continuously adapt to diverse domains. 
Specifically, it consists of DocVQA~\citep{mathew2021docvqa}, ChartQA~\citep{masry2022chartqa}, IconQA~\citep{lu2021iconqa}, and MedicalQA~\citep{he2020pathvqa}, which correspond to the document, chart, icon, and medical domains, respectively.
Adapt benchmark consists of visual instruction tuning datasets, such as M3IT~\citep{li2023m}, MiniGPT4~\citep{zhu2023minigpt}, MANTIS~\citep{jiang2024mantis}, LaMM~\citep{yin2023lamm}, and VisionFLAN~\citep{XuSH23}.
\oursbenchmark comprises NLVR2~\citep{Suhr2017ACO}, Bongard-OpenWorld~\citep{wu2024bongard}, Bongard-HOI~\citep{jiang2022bongard}, Co-Instruct-DB~\citep{wu2024towards}, DVQA~\citep{kafle2018dvqa}, HQ-Edit~\citep{hui2025hqedit}, and PatternCom~\citep{psomas2024composed}.
We summarize the task configuration of each benchmark, including \oursbenchmark in Tab.~\ref{tab:cil_setup}.

\begin{table*}[ht!]
  \centering
  \resizebox{1.0\linewidth}{!}{
    \begin{tabular}{lccc}
    \toprule
    Dataset & $\#$ of samples / task & $\#$ of tasks & Task order \\
    \cmidrule(lr){1-1} \cmidrule(lr){2-2} \cmidrule(lr){3-3} \cmidrule(lr){4-4} 
    
    COAST~\citep{cao2024continual} & 20,000 & 4 & ChartQA $\rightarrow$  DocVQA $\rightarrow$  IconQA $\rightarrow$  MedicalQA \\
    Adapt~\citep{maharana2025adaptinfty} & 80,000 & 4 & M3IT $\rightarrow$ MANTIS $\rightarrow$ LaMM $\rightarrow$ VisionFLAN \\
    \multirow{2}{*}{\oursbenchmark (Ours)} & \multirow{2}{*}{\makecell{50K / 30K / 100K / \\ 46K / 17K / 25K / 79K}} & 
    % \multirow{2}{*}{\oursbenchmark (Ours)} & \multirow{2}{*}{Imbalanced - Total 347K} & 
    \multirow{2}{*}{7} & Bongard-OpenWorld $\rightarrow$ NLVR2 $\rightarrow$  Co-Instruct-DB \\ 
    & & &  Bongard-HOI $\rightarrow$  PatternCom $\rightarrow$ DVQA $\rightarrow$ HQ Edit \\
    % SuperNI~\citep{wang2022super} & 1,000 & 15 & 
    Long Sequence~\citep{razdaibiedina2023progressive} & 10,000 & 15 & 
    DBPedia $\rightarrow$ RTE $\rightarrow$ WiC $\rightarrow$ Amazon $\rightarrow$ Yahoo \\ 
    & & & $\rightarrow$ AG News $\rightarrow$ MNLI $\rightarrow$ IMDB $\rightarrow$ MultiRC $\rightarrow$ CB \\ 
    & & & $\rightarrow$ Yelp $\rightarrow$ COPA $\rightarrow$ QQP $\rightarrow$ SST2 $\rightarrow$ BoolQ \\
    % Task2 $\rightarrow$ Task1290 $\rightarrow$ Task639 $\rightarrow$ Task1572 $\rightarrow$ Task1687 $\rightarrow$ Task181 \\
    % & & & $\rightarrow$ Task1729 $\rightarrow$ Task511 $\rightarrow$ Task1510 $\rightarrow$ Task591 $\rightarrow$ Task1590 \\
    % & & & $\rightarrow$ Task875 $\rightarrow$ Task363 $\rightarrow$ Task73 $\rightarrow$ Task748 \\
    TRACE~\citep{wang2023trace} & 5,000 & 8 & 
    NumGLUE-ds $\rightarrow$ ScienceQA $\rightarrow$ Py150 $\rightarrow$ C-STANCE \\
    & & & $\rightarrow$ FOMC $\rightarrow$ 20Minuten $\rightarrow$ NumGLUE-cm $\rightarrow$ MeetingBank \\
    \bottomrule 
    \end{tabular}
    }
  \vspace{-0.3em}
  \caption{
      \textbf{Task configurations of CIT benchmarks.}
  }
  \vspace{-.5em}
  \label{tab:cil_setup}
\end{table*}

\subsection{Sample Selection Baselines}
\label{sec:appx:baselines}

\paragraph{Self-Sup~\citep{sorscher2022beyond}.} 
Self-Sup performs K-means clustering on the output embeddings from the final layer of the pretrained model and selects the samples closest to each cluster centroid.
% By selecting samples that are closest to their respective cluster centroids, Self-Sup select the most representative sample in the underlying data distribution.
By doing so, the most representative sample in the underlying data distribution can be selected.
% This method tries to prioritize representativeness by capturing the central tendency within the embedding space.

\paragraph{GradNorm~\citep{katharopoulos2018not}.} GradNorm is an importance sampling method that prioritizes the selection of highly informative samples based on gradient magnitude. It computes each sample's gradient norm with respect to the model's last layer parameters and assigns higher selection probabilities to samples with larger gradient norms.

\paragraph{COINCIDE~\citep{lee2024concept}.}
COINCIDE leverages multi-layer output representations for K-means clustering. 
Specifically, features are extracted from five distinct layers of the MLLM to 
% form a comprehensive 
obtain the
representation of the candidate samples. 
Then it applies K-means clustering to these features, and computes 
% two metrics (\ie, transferability and density for each cluster) 
transferability and density metrics of each cluster
to determine how many samples should be selected from the respective cluster. 
Finally, within each cluster, 
% samples that best reflect the overall cluster distribution are chosen through computing maximum mean discrepancy (MMD)~\citep{kim2016examples} using greedy search.
the maximum mean discrepancy (MMD)~\citep{kim2016examples} is computed greedily to obtain the samples that best reflect the overall cluster distribution.

\paragraph{DBP~\citep{abbas2024effective}.} DBP aims to improve dataset quality by removing redundant and irrelevant samples. First, following Semdedup~\citep{abbas2023semdedup}, DBP applies K-means clustering to group similar samples with last layer outputs and eliminate semantic duplicates by retaining only the sample farthest from the cluster centroid among highly similar pairs. Next, DBP performs CLIP score filtering, removing samples with low text-image similarities. Then, 
% to remove low-quality samples, 
to prioritize more informative data, each cluster is evaluated for its complexity using two metrics: \( d_{\text{inter}} \), the distance from the cluster centroid to other centroids, and \( d_{\text{intra}} \), the average cosine distance of samples to their own centroid. A complexity score \( C_j = d_{\text{inter},j} \cdot d_{\text{intra},j} \) is calculated for each cluster \( j \), and samples are selected proportionally based on this score. Within each selected cluster, samples with the highest entropy are retained, ensuring that the final dataset is both diverse and rich in information.

\paragraph{InfoBatch~\citep{qin2024infobatch}.} InfoBatch introduces a soft pruning strategy to improve training efficiency while preserving learning dynamics. Unlike traditional hard pruning methods that permanently discard data and risk introducing bias, InfoBatch probabilistically excludes a subset of well-learned samples—identified by their low loss values—during each training epoch. Furthermore, to maintain the integrity of the training trajectory, the gradients of the remaining samples are rescaled such that the expected aggregated gradient approximates that of the full batch.

% \paragraph{Maxloss~\citep{}} Maxloss proposes a rank-based batch selection strategy in which training samples are sorted in descending order based on their loss values. The probability of a sample being selected decreases exponentially with its rank. Specifically, the top-ranked sample has the highest selection probability \( p_1 \), while the lowest-ranked sample has a probability \( p_N = p_1 / s \), where \( s \) denotes the ratio between the greatest and smallest selection probabilities.

\paragraph{DivBS~\citep{abbas2024effective}.} From each online batch, DivBS focuses on selecting a fixed subset of training data that is both informative and diverse by maximizing the orthogonalized representativeness. To do so, DivBS utilizes the last layer gradient. DivBS aims to explicitly reduce inter-sample redundancy, ensuring the retained samples capture complementary and non-overlapping aspects of the batch distribution.

\paragraph{TIVE~\citep{liu2024less}.} TIVE selects samples based on two criteria: instance influence and task difficulty, both derived from gradients computed across all layers of a reference model. Instance influence quantifies how much
% training
a given sample contributes to other samples during training, while task difficulty estimates the inherent complexity of learning a given task. Guided by these two metrics, TIVE prioritizes samples that are both highly influential and associated with difficult tasks.

\paragraph{Adapt-$\infty$~\citep{maharana2025adaptinfty}.} Rather than utilizing layer output representations, Adapt-$\infty$ uses gradients from the middle layer to cluster data into pseudo-tasks, enabling the model to identify and group related skills without requiring explicit task labels. It then performs multi-way sample selection within each cluster using a pool of scoring functions—such as entropy and the proposed Image Grounding score—to retain the most informative samples. Unlike our online continual instruction tuning setting, Adapt-$\infty$ was introduced in an offline continual instruction tuning setting, where task boundary assumptions exist,
limiting its real-world applicability.

\subsection{Experiments with Qwen-VL-2.5}
\label{sec:appx:qwen_results}

\begin{table*}[ht!]
  % \label{tab:headings}
  \centering
  \resizebox{\linewidth}{!}{
    \begin{tabular}{lcccccc}
    \toprule
    \multirow{3}{*}{Method} & \multicolumn{6}{c}{Selection Ratio (\%)} \\     \cmidrule(lr){2-7} 
    & \multicolumn{2}{c}{6.25} & \multicolumn{2}{c}{12.5} & \multicolumn{2}{c}{25.0} \\
    
    & $A_{avg} \ \uparrow$ & $A_{last} \ \uparrow$ & $A_{avg} \ \uparrow$ & $A_{last} \ \uparrow$ & $A_{avg} \ \uparrow$ & $A_{last} \ \uparrow$ \\ 

    \cmidrule(lr){1-1} \cmidrule(lr){2-3} \cmidrule(lr){4-5} \cmidrule(lr){6-7} 
    Full-Data Training & 34.59$\pm$0.44 &  41.90$\pm$.0.48 
    & 34.59$\pm$0.44 &  41.90$\pm$.0.48
    & 34.59$\pm$0.44 &  41.90$\pm$.0.48 \\

    \cmidrule(lr){1-1} \cmidrule(lr){2-3} \cmidrule(lr){4-5} \cmidrule(lr){6-7} 

    % Full-Training & 
    % \multicolumn{6}{c}{ $\pm$ } \\

    % \cmidrule(lr){1-1} \cmidrule(lr){2-3} \cmidrule(lr){4-5} \cmidrule(lr){6-7} 

    Random & 
    24.53$\pm$0.47 & 31.59$\pm$0.26 & 26.39$\pm$0.54 & 
    34.14$\pm$0.55 & 27.89$\pm$0.15 & 37.32$\pm$0.28 \\

    GradNorm {\footnotesize \color{blue}{(ICML 2018)}} & 
    23.85$\pm$0.39 & 29.69$\pm$0.84 & 25.96$\pm$0.57 & 
    33.28$\pm$0.82 & 26.47$\pm$0.19 & 35.13$\pm$1.14 \\

    Self-Sup {\footnotesize \color{blue}{(NeurIPS 2022)}} & 
    22.44$\pm$0.18 & 30.28$\pm$0.80 & 24.18$\pm$1.26 & 
    32.53$\pm$0.42 & 25.81$\pm$0.05 & 33.92$\pm$0.59 \\
    
    COINCIDE {\footnotesize \color{blue}{(EMNLP 2024)}} & 
    21.93$\pm$0.45 & 30.73$\pm$0.16 & 24.74$\pm$0.74 & 
    33.84$\pm$0.43 & 27.48$\pm$0.57 & 35.27$\pm$0.65 \\
    
    DBP {\footnotesize \color{blue}{(ICLR 2024)}} & 
    22.58$\pm$0.29 & 27.92$\pm$0.27 & 24.01$\pm$0.48 & 
    32.43$\pm$0.58 & 25.35$\pm$0.52 & 34.79$\pm$0.61 \\

    InfoBatch {\footnotesize \color{blue}{(ICLR 2024)}} & 
    24.60$\pm$0.36 & 32.88$\pm$0.54 & 27.48$\pm$0.39 & 
    \underline{34.26$\pm$1.39} & 28.93$\pm$0.41 & 38.12$\pm$0.42 \\

    DivBS {\footnotesize \color{blue}{(ICML 2024)}} & 
    25.45$\pm$0.64 & 33.18$\pm$0.51 & 28.29$\pm$0.48 & 
    \underline{36.08$\pm$0.85} & 30.82$\pm$0.28 & \textbf{39.87}$\pm$\textbf{0.74} \\

    TIVE {\footnotesize \color{blue}{(arXiv:2403)}} & 
    22.52$\pm$0.82 & 28.90$\pm$0.33 & 25.83$\pm$0.62 & 
    32.35$\pm$0.93 & 27.14$\pm$0.45 & 34.42$\pm$0.14 \\

    Adapt-$\infty$ {\footnotesize \textcolor{blue}{(ICLR 2025)}} &
    23.84$\pm$0.53 & 29.48$\pm$0.29 & 24.85$\pm$0.84 & 
    33.43$\pm$0.79 & 26.78$\pm$0.63 & 35.38$\pm$0.38 \\

    \cmidrule(lr){1-1} \cmidrule(lr){2-3} \cmidrule(lr){4-5} \cmidrule(lr){6-7} 
    
    \ours (\textbf{Ours}) & 
    \textbf{27.29}$\pm$\textbf{0.18} & \textbf{35.68}$\pm$\textbf{0.62} & \textbf{29.16}$\pm$\textbf{0.13} & \textbf{36.27}$\pm$\textbf{0.75} & 
    \textbf{31.04}$\pm$\textbf{0.38} & \underline{39.38$\pm$0.84} \\

    \bottomrule
    \end{tabular}
    }
    \caption{
    \textbf{Quantitative comparison between online sample selection methods on COAST benchmark.}
    % We use Qwen-VL-2.5 as an MLLM.
    % \ours outperforms the baselines across various selection ratios.
    % Bold indicates the highest performance, and underlined results are within the 0.05 significance level of the t-test.
    We use Qwen-VL-2.5-7B as the MLLM.
    Bold indicates the highest performance; underlined results are within the 0.05 t-test significance level. 
    'Full-Data Training' uses all data without selection.
    }
  \label{tab:main_qwen_coast}
  \vspace{-.6em}
\end{table*}

\begin{table*}[ht!]
  % \label{tab:headings}
  \centering
  \resizebox{\linewidth}{!}{
    \begin{tabular}{lcccccc}
    \toprule
    \multirow{3}{*}{Method} & \multicolumn{6}{c}{Selection Ratio (\%)} \\     \cmidrule(lr){2-7} 
    & \multicolumn{2}{c}{6.25} & \multicolumn{2}{c}{12.5} & \multicolumn{2}{c}{25.0} \\
    
    & $A_{avg} \ \uparrow$ & $A_{last} \ \uparrow$ & $A_{avg} \ \uparrow$ & $A_{last} \ \uparrow$ & $A_{avg} \ \uparrow$ & $A_{last} \ \uparrow$ \\ 

    \cmidrule(lr){1-1} \cmidrule(lr){2-3} \cmidrule(lr){4-5} \cmidrule(lr){6-7} 
    Full-Data Training & 
    45.73$\pm$1.68 & 43.92$\pm$0.85 &  
    45.73$\pm$1.68 & 43.92$\pm$0.85 & 
    45.73$\pm$1.68 & 43.92$\pm$0.85  \\
    
    \cmidrule(lr){1-1} \cmidrule(lr){2-3} \cmidrule(lr){4-5} \cmidrule(lr){6-7} 

    % Full-Training & 
    % \multicolumn{6}{c}{ $\pm$ } \\

    % \cmidrule(lr){1-1} \cmidrule(lr){2-3} \cmidrule(lr){4-5} \cmidrule(lr){6-7} 

    Random & 
    34.87$\pm$0.50 & 27.49$\pm$0.25 & 36.02$\pm$0.35 & 
    32.16$\pm$0.62 & 38.73$\pm$0.44 & 34.14$\pm$0.57 \\

    GradNorm {\footnotesize \color{blue}{(ICML 2018)}} & 
    32.53$\pm$0.36 & 25.41$\pm$0.72 & 34.70$\pm$0.28 & 
    30.25$\pm$0.43 & 35.38$\pm$0.40 & 32.26$\pm$0.42 \\

    Self-Sup {\footnotesize \color{blue}{(NeurIPS 2022)}} & 
    30.31$\pm$0.31 & 22.04$\pm$0.65 & 31.11$\pm$0.64 & 
    27.73$\pm$0.83 & 33.86$\pm$0.50 & 31.63$\pm$1.35 \\
    
    COINCIDE {\footnotesize \color{blue}{(EMNLP 2024)}} & 
    32.76$\pm$0.17 & 24.39$\pm$0.31 & 33.93$\pm$0.72 & 
    29.34$\pm$0.47 & 35.82$\pm$0.08 & 32.38$\pm$0.55 \\
    
    DBP {\footnotesize \color{blue}{(ICLR 2024)}} & 
    29.82$\pm$0.52 & 22.26$\pm$0.36 & 31.16$\pm$0.27 & 
    28.04$\pm$0.14 & 34.24$\pm$0.61 & 30.05$\pm$0.72 \\

    InfoBatch {\footnotesize \color{blue}{(ICLR 2024)}} & 
    34.95$\pm$0.20 & 27.19$\pm$0.18 & 35.46$\pm$0.83 & 
    30.77$\pm$0.54 & 38.25$\pm$0.72 & 34.10$\pm$0.50 \\

    DivBS {\footnotesize \color{blue}{(ICML 2024)}} & 
    34.24$\pm$0.75 & 28.63$\pm$0.91 & 37.24$\pm$0.41 & 
    31.68$\pm$1.39 & \underline{40.32$\pm$0.53} & \underline{36.59$\pm$1.08} \\

    TIVE {\footnotesize \color{blue}{(arXiv:2403)}} & 
    31.46$\pm$0.69 & 23.74$\pm$0.87 & 32.05$\pm$0.06 & 
    28.13$\pm$0.28 & 35.49$\pm$0.69 & 31.90$\pm$0.35 \\

    Adapt-$\infty$ {\footnotesize \textcolor{blue}{(ICLR 2025)}} &
    33.61$\pm$0.83 & 24.93$\pm$0.42 & 34.00$\pm$0.57 & 
    28.51$\pm$1.54 & 35.36$\pm$0.82& 32.37$\pm$0.63 \\

    \cmidrule(lr){1-1} \cmidrule(lr){2-3} \cmidrule(lr){4-5} \cmidrule(lr){6-7} 
    
    \ours (\textbf{Ours}) & 
    \textbf{36.04}$\pm$\textbf{0.26} & \textbf{30.78}$\pm$\textbf{0.84} & \textbf{38.72}$\pm$\textbf{0.62}  & 
    \textbf{33.10}$\pm$\textbf{0.43} & \textbf{41.13}$\pm$\textbf{0.91} & \textbf{37.36}$\pm$ \textbf{0.55}\\

    \bottomrule
    \end{tabular}
    }
    \caption{
    \textbf{Quantitative comparison between online sample selection methods on Adapt benchmark.}
    We use Qwen-VL-2.5-7B as the MLLM.
    Bold indicates the highest performance; underlined results are within the 0.05 t-test significance level. 
    'Full-Data Training' uses all data without selection.
    }
  \label{tab:main_qwen_adapt}
  \vspace{-.6em}
\end{table*}

In addition to experiments with LLaVA-1.5-7B~\citep{liu2024improved} in Sec.~\ref{sec:quanti_main}, we also compare \ours with sample selection baselines using Qwen-VL-2.5-7B~\citep{bai2025qwen2} as the MLLM. 
Specifically, we compare \ours with baselines on COAST, Adapt, and MICVIT, and summarize the results in Tab.~\ref{tab:main_qwen_coast}, Tab.~\ref{tab:main_qwen_adapt}, Tab.~\ref{tab:qwen_ours_625_125} respectively.
As shown in the tables, \ours outperforms the baselines, consistent with the results observed using LLaVA-1.5-7B.

\begin{table*}[t!]
  \centering
  \resizebox{0.7\linewidth}{!}{
    \begin{tabular}{lcccc}
    \toprule
    \multirow{3}{*}{Method} & \multicolumn{4}{c}{Selection Ratio (\%)} \\     \cmidrule(lr){2-5} 
    & \multicolumn{2}{c}{6.25} & \multicolumn{2}{c}{12.5} \\
    & $A_{avg} \ \uparrow$ & $A_{last} \ \uparrow$ & $A_{avg} \ \uparrow$ & $A_{last} \ \uparrow$ \\ 
    
    \cmidrule(lr){1-1} \cmidrule(lr){2-3} \cmidrule(lr){4-5} 
    Full-Data Training & 
    72.31$\pm$0.42 & 78.18$\pm$0.75 & 
    72.31$\pm$0.42 & 78.18$\pm$0.75 \\

    \cmidrule(lr){1-1} \cmidrule(lr){2-3} \cmidrule(lr){4-5}

    Random & 
    59.23$\pm$0.85 & 65.94$\pm$0.38 & 
    62.18$\pm$0.19 & 69.03$\pm$0.10 \\

    GradNorm & 
    57.49$\pm$0.92 & 63.38$\pm$0.64 & 
    63.90$\pm$0.22 & 69.34$\pm$0.47 \\

    Self-Sup & 
    54.29$\pm$0.56 & 61.42$\pm$0.31 & 
    61.37$\pm$0.99 & 68.32$\pm$0.38 \\
    
    COINCIDE & 
    57.34$\pm$0.77 & 62.90$\pm$0.48 & 
    62.38$\pm$0.73 & 69.23$\pm$0.44 \\
    
    DBP & 
    54.53$\pm$0.43 & 60.38$\pm$0.56 & 
    60.63$\pm$0.30 & 67.81$\pm$0.52 \\

    InfoBatch & 
    58.14$\pm$0.28 & 65.38$\pm$0.87 & 
    64.04$\pm$0.42 & 68.26$\pm$1.31 \\

    DivBS & 
    \underline{60.28$\pm$0.98} & 67.48$\pm$0.29 & 
    \underline{65.26$\pm$1.34} & 70.48$\pm$0.38 \\

    TIVE & 
    55.84$\pm$0.49 & 63.37$\pm$0.83 & 
    62.11$\pm$0.45 & 67.42$\pm$0.73 \\

    Adapt-$\infty$ &
    56.32$\pm$0.97 & 62.48$\pm$0.38 & 
    62.35$\pm$0.49 & 67.82$\pm$1.24 \\

    \cmidrule(lr){1-1} \cmidrule(lr){2-3} \cmidrule(lr){4-5}
    
    \ours (\textbf{Ours}) & 
    \textbf{61.64}$\pm$\textbf{0.43} & \textbf{69.43}$\pm$\textbf{0.72} & 
    \textbf{66.54}$\pm$\textbf{0.88} & \textbf{72.35}$\pm$\textbf{0.28} \\
    \bottomrule
    \end{tabular}
  }
  \vspace{-0.5em}
  \caption{
  \textbf{Quantitative comparison with Qwen-VL-2.5-7B on \oursbenchmark benchmark under 6.25\% and 12.5\% selection ratios.}
  Bold indicates the highest performance; underlined results are within the 0.05 t-test significance level.
  }
  \label{tab:qwen_ours_625_125}
  \vspace{-1em}
\end{table*}
\subsection{Details on Determining \texorpdfstring{$I_T$}{I\_T}}
\label{sec:appx:threshold}

As described in Sec.~\ref{subsec:ours_one}, we probabilistically select a sample based on its relative informativeness $\hat{I}$, with the selection probability defined as:
\begin{equation}
p(\hat{I}) = \sigma(\hat{I} - I_{T}),
\end{equation}
where $\sigma(\cdot)$ denotes the sigmoid function and $I_T$ is a predefined threshold.
We assume $\hat{I} \sim \mathcal{N}(0, 1)$ 
% ,as we have shown in Fig.~\ref{fig:qqplot} 
that the original (\ie, pre-normalized) informativeness $I$ approximately follows a normal distribution.
Based on this normal distribution assumption, the expected selection rate across the distribution is:
\begin{align}
f(I_{T}) &= \mathbb{E}_{\hat{I}  \sim \mathcal{N}(0,1)}[\sigma(\hat{I}  - I_{T})] \\
&= \int_{-\infty}^{\infty} \sigma(\hat{I}  - I_{T}) \cdot \phi(\hat{I}) \, d\hat{I},
\end{align}
where $\phi(z)$ denotes the standard normal probability density function.

Given target selection rate $r \in [0,1]$ (\eg, 0.125), we find $I_{T}$ such that $f(I_{T}) = r$. 
Since the integral has no closed form, we approximate it numerically using a Riemann sum and solve:
\begin{equation}
\label{eq:I_t_determine}
I_{T} = \arg\min_{I_{T}} (f(I_{T}) - r)^2.
\end{equation}
Applying Eq.~\ref{eq:I_t_determine}, selection ratios $r$ of 0.0625, 0.125, and 0.25 yield corresponding $I_{T}$ values of 2.06, 1.53, and 0.89, respectively. 
However, data distribution variations and the probabilistic selection process may cause the actual proportion of selected samples to deviate from the target ratio. 
To ensure fair comparison with baseline methods, we fine-tune $I_{T}$ to achieve sample counts that closely approximate but remain marginally below the target selection ratio.

% often falls short of the intended selection ratio.

\subsection{Experiments on Various Model Scales}
\label{sec:appx:scale_exp}

In addition to LLaVA-1.5-7B and Qwen-VL-2.5-7B in the main paper, we compare baselines with various sizes of LLaVAs (\ie, LLaVA-1.5-1B, 3B, and 13B) in Tab.~\ref{tab:main_modelscale_micvit} and Tab.~\ref{tab:main_modelscale_coast} and Qwen-VL-2.5-0.5B in Tab.~\ref{tab:various_size_qwen}. 
As shown in the tables, \ours consistently outperforms the baselines across architectures and model sizes by a significant margin, demonstrating its general applicability.

\begin{table*}[t!]
  \centering
  \resizebox{\linewidth}{!}{
    \begin{tabular}{lcccccc}
    \toprule
    \multirow{3}{*}{Method} & \multicolumn{6}{c}{Model Scale} \\     
    \cmidrule(lr){2-7} 
    & \multicolumn{2}{c}{LLaVA-1.5-1B} & \multicolumn{2}{c}{LLaVA-1.5-3B} & \multicolumn{2}{c}{LLaVA-1.5-13B} \\
    & $A_{avg} \ \uparrow$ & $A_{last} \ \uparrow$ & $A_{avg} \ \uparrow$ & $A_{last} \ \uparrow$ & $A_{avg} \ \uparrow$ & $A_{last} \ \uparrow$ \\ 
    \cmidrule(lr){1-1} \cmidrule(lr){2-3} \cmidrule(lr){4-5} \cmidrule(lr){6-7} 

    Random & 
    55.04$\pm$0.12 & 59.75$\pm$0.30 & 
    57.37$\pm$0.12 & 60.39$\pm$0.68 & 
    \underline{67.49$\pm$0.47} & \underline{73.60$\pm$0.49} \\

    GradNorm {\footnotesize \color{blue}{(ICML 2018)}} & 
    53.44$\pm$0.05 & 57.39$\pm$0.22 & 
    56.19$\pm$0.70 & 60.73$\pm$0.08 & 
    65.32$\pm$0.53 & 71.50$\pm$0.98 \\

    Self-Sup {\footnotesize \color{blue}{(NeurIPS 2022)}} & 
    52.01$\pm$0.33 & 55.96$\pm$0.27 & 
    54.59$\pm$0.53 & 56.76$\pm$0.50 & 
    64.71$\pm$0.71 & 71.62$\pm$0.57 \\

    COINCIDE {\footnotesize \color{blue}{(EMNLP 2024)}} & 
    53.03$\pm$0.51 & 56.94$\pm$0.16 & 
    56.04$\pm$0.40 & 59.55$\pm$0.39 & 
    65.47$\pm$0.85 & 72.25$\pm$0.86 \\

    DBP {\footnotesize \color{blue}{(ICLR 2024)}} & 
    51.02$\pm$0.23 & 55.62$\pm$0.34 & 
    53.32$\pm$0.68 & 55.89$\pm$0.39 & 
    62.27$\pm$0.02 & 68.37$\pm$0.02 \\

    InfoBatch {\footnotesize \color{blue}{(ICLR 2024)}} & 
    55.48$\pm$0.08 & \underline{60.29$\pm$1.09} & 
    \underline{58.92$\pm$1.31} & 60.39$\pm$0.64 & 
    66.78$\pm$0.45 & 72.04$\pm$0.93 \\

    DivBS {\footnotesize \color{blue}{(ICML 2024)}} & 
    57.24$\pm$0.12 & 60.22$\pm$0.42 & 
    \underline{58.02$\pm$0.72} & \underline{62.15$\pm$0.52} & 
    \underline{67.28$\pm$0.27} & 73.63$\pm$0.18 \\

    TIVE {\footnotesize \color{blue}{(arXiv:2403)}} & 
    51.34$\pm$0.11 & 56.26$\pm$0.08 & 
    54.82$\pm$0.12 & 58.90$\pm$0.37 & 
    64.74$\pm$0.02 & 71.50$\pm$0.02 \\

    Adapt-$\infty$ {\footnotesize \textcolor{blue}{(ICLR 2025)}} &
    54.09$\pm$0.15 & 57.09$\pm$0.21 & 
    55.29$\pm$0.45 & 56.81$\pm$0.69 & 
    63.69$\pm$0.68 & 70.46$\pm$0.81 \\

    \cmidrule(lr){1-1} \cmidrule(lr){2-3} \cmidrule(lr){4-5} \cmidrule(lr){6-7} 

    \ours (\textbf{Ours}) & 
    \textbf{57.86}$\pm$\textbf{0.10} & \textbf{62.17}$\pm$\textbf{0.25} & 
    \textbf{60.46}$\pm$\textbf{0.26} & \textbf{64.01}$\pm$\textbf{0.40} & 
    \textbf{68.04}$\pm$\textbf{0.21} & \textbf{74.60}$\pm$\textbf{0.22} \\

    \bottomrule
    \end{tabular}
    }
    \vspace{-0.5em}
    \caption{
    \textbf{Quantitative comparison between online sample selection methods across model scales on \oursbenchmark under selection ratio 6.25\%.}
    Bold indicates the highest performance; underlined results are within the 0.05 t-test significance level. 
    }
  \label{tab:main_modelscale_micvit}
  \vspace{-.6em}
\end{table*}

\begin{table*}[t!]
  \centering
  \resizebox{\linewidth}{!}{
    \begin{tabular}{lcccccc}
    \toprule
    \multirow{3}{*}{Method} & \multicolumn{6}{c}{Model Scale} \\     
    \cmidrule(lr){2-7} 
    & \multicolumn{2}{c}{LLaVA-1.5-1B} & \multicolumn{2}{c}{LLaVA-1.5-3B} & \multicolumn{2}{c}{LLaVA-1.5-13B} \\
    & $A_{avg} \ \uparrow$ & $A_{last} \ \uparrow$ & $A_{avg} \ \uparrow$ & $A_{last} \ \uparrow$ & $A_{avg} \ \uparrow$ & $A_{last} \ \uparrow$ \\ 
    \cmidrule(lr){1-1} \cmidrule(lr){2-3} \cmidrule(lr){4-5} \cmidrule(lr){6-7} 

    Random & 
    20.05$\pm$0.21 & 28.04$\pm$0.15 & 
    21.20$\pm$0.56 & 29.30$\pm$0.22 & 
    \underline{25.39$\pm$0.90} & 33.84$\pm$0.72 \\

    GradNorm {\footnotesize \color{blue}{(ICML 2018)}} & 
    19.13$\pm$0.33 & 27.49$\pm$0.11 & 
    19.95$\pm$0.80 & 28.73$\pm$0.39 & 
    23.38$\pm$0.56 & 32.03$\pm$0.51 \\

    Self-Sup {\footnotesize \color{blue}{(NeurIPS 2022)}} & 
    18.17$\pm$0.41 & 26.36$\pm$0.20 & 
    17.49$\pm$0.86 & 24.76$\pm$0.80 & 
    21.47$\pm$0.37 & 29.35$\pm$0.22 \\

    COINCIDE {\footnotesize \color{blue}{(EMNLP 2024)}} & 
    19.46$\pm$1.17 & 28.27$\pm$0.09 & 
    20.11$\pm$0.16 & 28.74$\pm$0.90 & 
    23.77$\pm$0.79 & 31.74$\pm$0.27 \\

    DBP {\footnotesize \color{blue}{(ICLR 2024)}} & 
    17.98$\pm$0.14 & 25.57$\pm$0.06 & 
    17.05$\pm$0.69 & 25.71$\pm$1.83 & 
    22.24$\pm$0.22 & 31.95$\pm$0.62 \\

    InfoBatch {\footnotesize \color{blue}{(ICLR 2024)}} & 
    18.36$\pm$0.18 & 25.47$\pm$0.04 & 
    19.92$\pm$0.89 & 29.27$\pm$0.56 & 
    24.05$\pm$0.73 & 30.48$\pm$0.64 \\

    DivBS {\footnotesize \color{blue}{(ICML 2024)}} & 
    21.21$\pm$0.24 & 28.52$\pm$0.13 & 
    21.39$\pm$0.35 & 31.41$\pm$0.80 & 
    \underline{26.22$\pm$0.10} & \underline{35.41$\pm$0.10} \\

    TIVE {\footnotesize \color{blue}{(arXiv:2403)}} & 
    18.04$\pm$0.25 & 26.35$\pm$0.30 & 
    17.47$\pm$0.74 & 26.11$\pm$0.62 & 
    \underline{24.58$\pm$0.34} & 32.86$\pm$0.60 \\

    Adapt-$\infty$ {\footnotesize \textcolor{blue}{(ICLR 2025)}} &
    19.91$\pm$0.16 & 26.81$\pm$0.28 & 
    18.69$\pm$0.61 & 28.00$\pm$0.58 & 
    22.52$\pm$0.61 & 31.20$\pm$0.21 \\

    \cmidrule(lr){1-1} \cmidrule(lr){2-3} \cmidrule(lr){4-5} \cmidrule(lr){6-7} 

    \ours (\textbf{Ours}) & 
    \textbf{23.03}$\pm$\textbf{0.18} & \textbf{30.37}$\pm$\textbf{0.12} & 
    \textbf{23.08}$\pm$\textbf{0.17} & \textbf{32.84}$\pm$\textbf{0.13} & 
    \textbf{26.52}$\pm$\textbf{0.52} & \textbf{36.81}$\pm$\textbf{0.65} \\

    \bottomrule
    \end{tabular}
    }
    \vspace{-0.5em}
    \caption{
    \textbf{Quantitative comparison between online sample selection methods across model scales on COAST benchmark under selection ratio 6.25\%.}
    Bold indicates the highest performance; underlined results are within the 0.05 t-test significance level. 
    }
  \label{tab:main_modelscale_coast}
  \vspace{-.6em}
\end{table*}

\begin{table}[t!]
  \centering
  \resizebox{\linewidth}{!}{
    \begin{tabular}{lcccc}
    \toprule
    \multirow{2}{*}{Method} & \multicolumn{2}{c}{\oursbenchmark} & \multicolumn{2}{c}{COAST} \\
    
    & $A_{avg} \ \uparrow$ & $A_{last} \ \uparrow$ & $A_{avg} \ \uparrow$ & $A_{last} \ \uparrow$ \\ 

    \cmidrule(lr){1-1} \cmidrule(lr){2-3} \cmidrule(lr){4-5} 

    Random & 
    55.66$\pm$0.36 & \underline{57.97$\pm$0.61} &
    \underline{17.64$\pm$0.50} & \underline{23.24$\pm$1.44} \\

    GradNorm & 
    54.39$\pm$1.59 & 56.90$\pm$0.41 &
    14.94$\pm$0.79 & 22.54$\pm$0.31 \\

    Self-Sup & 
    53.20$\pm$0.71 & 55.95$\pm$0.58 &
    15.22$\pm$0.64 & 19.77$\pm$0.30 \\

    COINCIDE & 
    55.18$\pm$0.16 & 58.32$\pm$0.61 &
    13.87$\pm$0.50 & 21.67$\pm$0.15 \\

    DBP & 
    53.43$\pm$0.63 & 57.25$\pm$1.50 &
    14.20$\pm$0.28 & 18.82$\pm$0.20 \\

    InfoBatch & 
    54.90$\pm$0.46 & 54.30$\pm$0.54 &
    \underline{16.74$\pm$0.52} & \underline{23.54$\pm$0.35} \\

    DivBS & 
    55.52$\pm$0.12 & \underline{57.22$\pm$0.17} &
    \underline{17.35$\pm$0.68} & \underline{23.57$\pm$0.15} \\

    TIVE & 
    52.15$\pm$1.06 & 54.12$\pm$0.59 &
    15.33$\pm$0.22 & \underline{22.75$\pm$1.30} \\

    Adapt-$\infty$ &
    54.89$\pm$0.53 & 55.65$\pm$0.91 &
    \underline{17.58$\pm$0.51} & 22.29$\pm$0.19 \\

    \cmidrule(lr){1-1} 
    \cmidrule(lr){2-3} \cmidrule(lr){4-5}
    
    OASIS (\textbf{Ours}) & 
    \textbf{57.66}$\pm$\textbf{0.17} & \textbf{59.84}$\pm$\textbf{0.64} &
    \textbf{19.02}$\pm$\textbf{0.71} & \textbf{25.08}$\pm$\textbf{0.59} \\

    \bottomrule
    \end{tabular}
    }
    \vspace{-0.5em}
    \caption{
    \textbf{Quantitative comparison between online sample selection methods on \oursbenchmark and COAST benchmark under selection ratio 6.25\% at QwenVL-0.5B.}
    Bold indicates the highest performance; underlined results are within the 0.05 t-test significance level. 
    }
  \label{tab:various_size_qwen}
  \vspace{-.6em}
\end{table}

\subsection{Detailed Experiments on Fast Adaptation}
\label{sec:appx:appen_few-shot}

\begin{figure*}[h]
    \vspace{-.5em}
    \centering   
    \includegraphics[width=\linewidth]{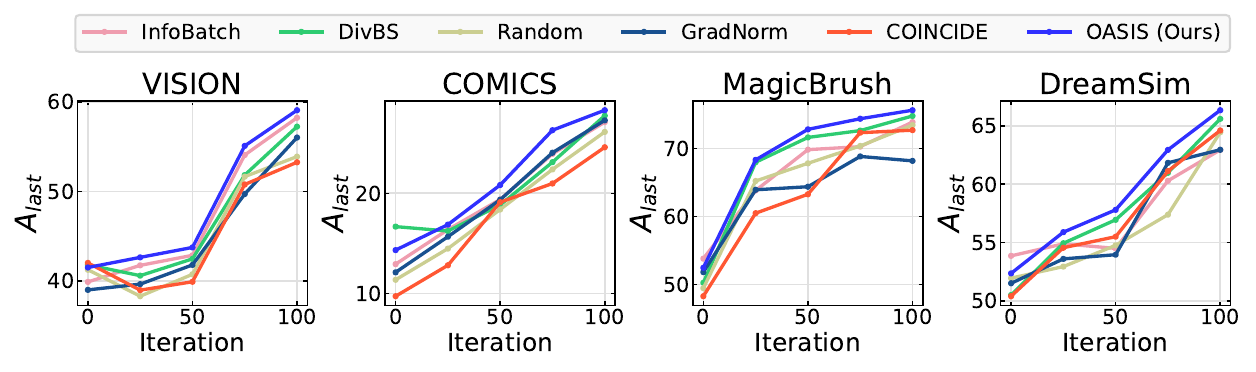}
    
    % {\vspace{-0.5em} \footnotesize (a) \oursbenchmark}
    
    % \centering   
    % \includegraphics[width=\linewidth]{figures/fewshot_coast.pdf}
    
    % {\vspace{-0.5em} \footnotesize (b) COAST}
    \vspace{-0.5em}
    \caption{
    \textbf{Comparison of fast adaptation performance.}
    After CIT of LLaVA-1.5-7B on subsets (25\% of the full data), selected using each sample selection baseline from COAST, we fine-tune the model for 100 epochs on each downstream task (\ie, VISION, COMICS, MagicBrush, and DreamSim). 
    }
    \label{fig:fewshot_coast}
\end{figure*}

Beyond the fast adaptation results on \oursbenchmark (Sec.~\ref{sec:quanti_main}), we also evaluate fast adaptation performance on COAST. 
As shown in Fig.~\ref{fig:fewshot_coast}, subsets selected by \ours consistently achieve superior results across all downstream tasks, highlighting their strong generalizability. 
While zero-shot performance (iteration 0) may vary depending on task–subset similarity, fine-tuning results show that \ours enables faster adaptation, confirming the superior generalization of models trained on its selected subsets.

\subsection{Different Information Metrics}
\label{sec:appx:info_exp}
We compare \ours using Fisher Information to other metrics for measuring the informativeness $I$ of samples, including Entropy~\citep{coleman2020selection}, Perplexity~\citep{li2023quantity}, and EL2N~\citep{paul2021deep}.
% We apply existing metrics for calculating how informative or hard each sample is to train (\eg, Entropy\citep{coleman2020selection}, Perplexity\citep{li2023quantity}), and EL2N~\citep{paul2021deep}) in place of Fisher Information (FI) in \ours. 
As shown in Tab.~\ref{tab:info_exp}, using sample-wise FI as our Informativeness outperforms other metrics.

\begin{table}[ht]
  \centering
  \resizebox{\linewidth}{!}{
    \begin{tabular}{lcccc}
    \toprule
    \multirow{2}{*}{Method} & \multicolumn{2}{c}{\oursbenchmark} & \multicolumn{2}{c}{COAST} \\
   
    & $A_{\text{last}} \ \uparrow$ & $A_{\text{avg}} \ \uparrow$ & $A_{\text{last}} \ \uparrow$ & $A_{\text{avg}} \ \uparrow$ \\
    
    \cmidrule(lr){1-1} 
    \cmidrule(lr){2-3} \cmidrule(lr){4-5}

    Entropy & \underline{62.75$\pm$0.84} & \underline{70.08$\pm$0.99} & 22.53$\pm$0.46 & 30.42$\pm$0.57 \\
    
    Perplexity & 59.86$\pm$1.37 & 64.49$\pm$0.72 & 20.01$\pm$0.73 & 28.64$\pm$0.82 \\
    
    EL2N & 60.36$\pm$1.44 & 67.15$\pm$0.81 & 23.28$\pm$0.56 & 31.81$\pm$0.60 \\

    \cmidrule(lr){1-1} 
    \cmidrule(lr){2-3} \cmidrule(lr){4-5}
    
    FI (Ours) & \textbf{64.39}$\pm$\textbf{0.58} & \textbf{71.26}$\pm$\textbf{0.72} & \textbf{25.67}$\pm$\textbf{0.35} & \textbf{34.23}$\pm$\textbf{0.38} \\

    \bottomrule
    \end{tabular}
  }
  \caption{
    \textbf{Comparison of different information metrics.} On \oursbenchmark and COAST benchmarks, we apply different existing information metrics in place of Fisher Information (FI) to \ours, when selection ratio is 6.25\%. Bold indicates the highest performance, and underlined results are within the 0.05 significance level of the t-test.
  }
  \label{tab:info_exp}
\end{table}

\subsection{Effect of Task Order}
\label{sec:appx:ablation_taskorder}
    
In Tab.~\ref{tab:task_order}, we examine the effect of task order using three different task sequences for the top-3 performing baselines and \ours on COAST and \oursbenchmark.
For each setting, 6.25\% of the training data is selected using \ours or the top three baseline methods.
% In each experiment, 6.25\% of the total training data is selected and trained using either \ours or the three best-performing selection method baselines. 
The results demonstrate that regardless of the change in task order, \ours outperforms other selection methods.
% The results demonstrate that while task order influences the final performance for each task, its impact on overall performance is minimal, with less than 1\% difference in last accuracy across different orders.

\begin{table}[t!]
  \centering
  \resizebox{\linewidth}{!}{
    \begin{tabular}{lllccc}
    \toprule
    
    \multirow{1}{*}{Benchmark} & \multirow{1}{*}{Order} & \multirow{1}{*}{Method} &\multirow{1}{*}{ $A_{avg}\uparrow$} & \multirow{1}{*}{$A_{last}\uparrow$}\\
    \cmidrule(lr){1-1} \cmidrule(lr){2-2} \cmidrule(lr){3-3} \cmidrule(lr){4-5}
    \multirow{14}{*}{\oursbenchmark} & \multirow{4}{*}{ONCHoPDH} 
    & Random & 61.15$\pm$0.34 & 67.29$\pm$0.61 \\
    & & InfoBatch & 60.82$\pm$0.75 & 68.88$\pm$1.09 \\
    & & DivBS & 61.07$\pm$0.58 & 69.06$\pm$0.54 \\
    \cmidrule(lr){3-3}  \cmidrule(lr){4-5}
    & & \ours (Ours) & \textbf{64.39}$\pm$\textbf{0.58} & \textbf{71.76}$\pm$\textbf{0.72} \\
    \cmidrule(lr){2-2} \cmidrule(lr){3-3} \cmidrule(lr){4-5}
    & \multirow{4}{*}{HDPHoCNO} 
    & Random & 57.85$\pm$0.63 & 66.94$\pm$0.74\\
    & & InfoBatch & 58.04$\pm$0.47 & 67.69$\pm$0.80 \\
    & & DIVBS & 58.39$\pm$0.28 & 69.32$\pm$0.53 \\
    \cmidrule(lr){3-3}  \cmidrule(lr){4-5}
    & & \ours (Ours) & \textbf{61.87}$\pm$\textbf{0.82} & \textbf{70.98}$\pm$\textbf{0.64} \\
    \cmidrule(lr){2-2} \cmidrule(lr){3-3} \cmidrule(lr){4-5}
    & \multirow{4}{*}{PNHOHoCD} 
    & Random & 60.31$\pm$0.88 & 68.18$\pm$0.62 \\
    & & InfoBatch & 60.25$\pm$0.35 & 67.52$\pm$0.16 \\
    & & DIVBS & 61.70$\pm$0.74 & 69.16$\pm$0.43 \\
    \cmidrule(lr){3-3}  \cmidrule(lr){4-5}
    & & \ours (Ours) & \textbf{63.12}$\pm$\textbf{0.58 }& \textbf{71.31}$\pm$\textbf{0.91} \\

    \midrule

    \multirow{14}{*}{COAST} & \multirow{4}{*}{cdim} 
    & Random & 23.57$\pm$0.17 & 30.80$\pm$0.30 \\
    & & InfoBatch & 22.93$\pm$0.73 & 29.14$\pm$0.56 \\
    & & DivBS & 23.41$\pm$0.14 & 31.72$\pm$0.18 \\
    \cmidrule(lr){3-3}  \cmidrule(lr){4-5}
    & & \ours (Ours) & \textbf{25.67}$\pm$\textbf{0.35} & \textbf{34.23}$\pm$\textbf{0.38} \\
    \cmidrule(lr){2-2} \cmidrule(lr){3-3} \cmidrule(lr){4-5}
    & \multirow{4}{*}{imcd}
    & Random & 21.45$\pm$0.39 & 29.25$\pm$0.45\\
    & & InfoBatch & 22.39$\pm$0.53 & 30.04$\pm$0.72 \\
    & & DIVBS & 23.84$\pm$0.26 & 31.10$\pm$0.56 \\
    \cmidrule(lr){3-3}  \cmidrule(lr){4-5}
    & & \ours (Ours) & \textbf{25.01}$\pm$\textbf{0.41} & \textbf{33.96}$\pm$\textbf{0.53} \\
    \cmidrule(lr){2-2} \cmidrule(lr){3-3} \cmidrule(lr){4-5}
    & \multirow{4}{*}{dmci} 
    & Random & 22.17$\pm$0.42 & 30.64$\pm$0.33\\
    & & InfoBatch & 20.31$\pm$0.25 & 29.53$\pm$1.28\\
    & & DIVBS & 22.92$\pm$0.33 & \underline{32.31$\pm$0.74}\\
    \cmidrule(lr){3-3}  \cmidrule(lr){4-5}
    & & \ours (Ours) & \textbf{24.36}$\pm$\textbf{0.59} & \textbf{33.28}$\pm$\textbf{0.26} \\
    
    \bottomrule 
    \end{tabular}
    }
  \vspace{-.3em}
  \caption{
      \textbf{Ablation on task order on \oursbenchmark and COAST benchmarks.} 
      In \oursbenchmark, ONCHoPDH, HDPHoCNO, and PNHOHoCD denote task orders of Bongard-OpenWorld $\rightarrow$ NLVR2 $\rightarrow$ Co-Instruct-DB, Bongard-HOI $\rightarrow$ PatternCom $\rightarrow$ DVQA $\rightarrow$ HQ Edit,  HQ Edit $\rightarrow$ DVQA $\rightarrow$ PatternCom $\rightarrow$ Bongard-HOI $\rightarrow$ NLVR2 $\rightarrow$ Bongard-OpenWorld, and PatternCom $\rightarrow$ NLVR2 $\rightarrow$ HQ Edit $\rightarrow$ Bongard-Openworld $\rightarrow$ Bongard-HOI $\rightarrow$ Co-Instruct-DB $\rightarrow$ DVQA, respectively.
      In COAST, cdim, imcd, and dmci denote task orders of ChartQA $\rightarrow$ DocVQA $\rightarrow$ IconQA $\rightarrow$ MedicalQA, IconQA $\rightarrow$ MedicalQA $\rightarrow$ ChartQA $\rightarrow$ DocVQA, and DocVQA $\rightarrow$ MedicalQA $\rightarrow$ ChartQA $\rightarrow$ IconQA, respectively.
  }
  \vspace{-.5em}
  \label{tab:task_order}
\end{table}

\subsection{Effect of EMA Ratio \texorpdfstring{$\beta$}{beta}}
\label{sec:appx:ema_ratio}
We further evaluate the effect of EMA ratio $\beta$ on \ours, and summarize the results in Tab.~\ref{tab:ema_ablation}.
As shown, extremely large or small values of 
$\beta$ degrade performance: high 
$\beta$ fails to capture the informativeness of past batches, while low 
$\beta$ overemphasizes outdated informativeness. 
To balance this trade-off, we use a moderate value of $\beta=0.9$ across all benchmarks and selection ratios.

\begin{table}[ht!]
  \centering
  \resizebox{0.8\linewidth}{!}{
    \begin{tabular}{lcc}
    \toprule
    % & \multicolumn{2}{c}{Overall Results} \\
    EMA Ratio $\beta$ & $A_{avg}\uparrow$ & $A_{last}\uparrow$ \\
    \cmidrule(lr){1-1} \cmidrule(lr){2-3} 
    0.7 & 22.24$\pm$0.25 & 31.93$\pm$0.75 \\
    0.9 & \textbf{24.36}$\pm$\textbf{0.59} & \textbf{33.28}$\pm$\textbf{0.26} \\
    0.99 & 22.80$\pm$0.42 & 32.73$\pm$0.64 \\
    0.999 & 23.05$\pm$0.57 & 32.94$\pm$0.81 \\
    \bottomrule 
    \end{tabular}
    }
  \vspace{-.3em}
  \caption{
      \textbf{Ablation on EMA ratio.}
  }
  \vspace{-.5em}
  \label{tab:ema_ablation}
\end{table}

\subsection{Accuracy Over Time}
\label{sec:appx:avg_accuracy}
\begin{figure*}[t!]
    % \vspace{-.6em}
    \centering   
    \includegraphics[width=\linewidth]{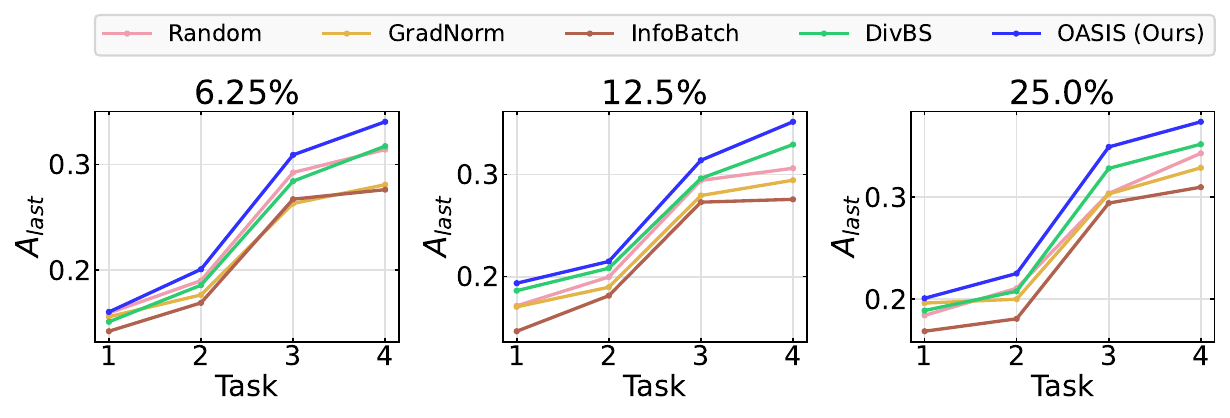}
    \vspace{-1.5em}
    \caption{
    \textbf{Average accuracies over time on COAST benchmark under different selection ratios.}
    Performance at task $t$ denotes the average accuracy over all seen tasks up to that point (\ie, task 1 through task $t$).
    We use LLaVA-1.5-7B as the model across all selection ratios.
    }
    \label{fig:taskacc_ratio_change}
    \vspace{-1em}
\end{figure*}

\begin{figure*}[t!]
    % \vspace{-.6em}
    \centering   
    \includegraphics[width=\linewidth]{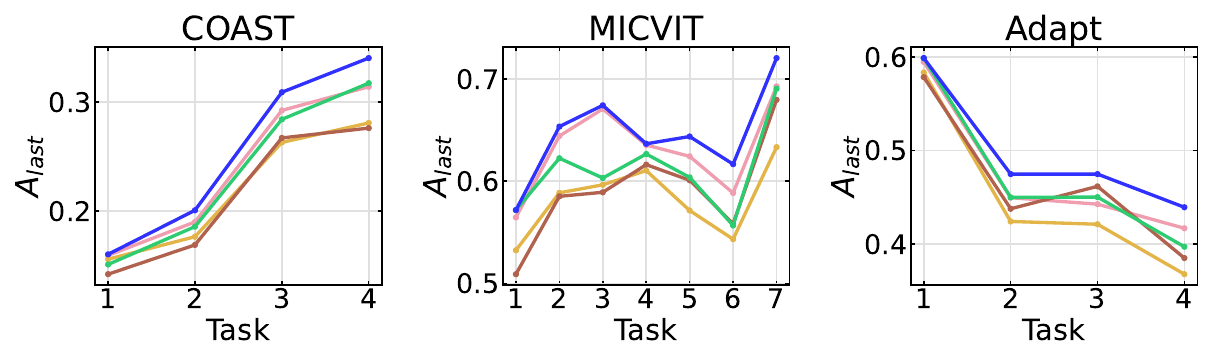}
    \caption{
    \textbf{Average accuracy over time across benchmarks.}
    Performance at task $t$ denotes the average accuracy over all seen tasks up to that point (\ie, task 1 through task $t$).
    We use a selection ratio of 6.25\% across all benchmarks.
    We use LLaVA-1.5-7B as the model.
    }
    \label{fig:taskacc_benchmark_change}
    \vspace{-1.5em}
\end{figure*}

We compare the average accuracy of seen tasks at several training time points between our method and baselines.
Specifically, we evaluate performance across different selection ratios and benchmarks, with results shown in Fig.\ref{fig:taskacc_ratio_change} and Fig.\ref{fig:taskacc_benchmark_change}, respectively.
As shown in the figures, our method outperforms baselines at all time points, demonstrating that its superior performance is not limited to specific intervals but is consistently maintained throughout the entire CIT process.

\subsection{Comparison of Computational Costs}
\label{sec:appx:cost_comparison}

We compare the computational cost of \ours with selection baselines in Tab.~\ref{tab:comp_cost}.
As shown, all methods require forward passes over candidate samples.
Sample selection methods (\eg, Self-Sup and COINCIDE) that rely solely on features do not incur additional backward pass costs.
In contrast, gradient-based methods (\eg, \ours and GradNorm) involve backward passes to compute sample-wise gradients.
However, unlike TIVE and Adapt-$\infty$, which introduce substantial overhead due to full-layer gradient computations or intermediate layer calculations, \ours leverages only the last-layer gradients, resulting in negligible overhead compared to the forward pass (\ie, less than 3.5\%).

\begin{table*}[ht!]
    \centering
    \resizebox{\linewidth}{!}{
    {
    \begin{tabular}{ccc}
    \toprule
    \multirow{1}{*}{Methods} & Overhead Type & Relative $\mathcal{C}$ to \ours \\
    \cmidrule(lr){1-1} \cmidrule(lr){2-2} \cmidrule(lr){3-3} 
    GradNorm {\footnotesize \color{blue}{(ICML 2018)}}  & Forward Pass + Last Layer Gradient Compute & 1.000 \\
    Self-Sup {\footnotesize \color{blue}{(NeurIPS 2022)}} & Forward Pass & 0.976\\
    COINCIDE {\footnotesize \color{blue}{(EMNLP 2024)}}  & Forward Pass & 0.976 \\
    DBP {\footnotesize \color{blue}{(ICLR 2024)}} & Forward Pass & 0.976 \\
    InfoBatch {\footnotesize \color{blue}{(ICLR 2024)}} & Forward Pass & 0.976 \\
    DivBS {\footnotesize \color{blue}{(ICML 2024)}} & Forward Pass + Last Layer Gradient Compute & 1.000 \\
    TIVE {\footnotesize \textcolor{blue}{(arXiv:2403)}} & Forward Pass + Full Layer Gradient Compute & 2.038 \\
    Adapt-$\infty$ {\footnotesize \textcolor{blue}{(ICLR 2025)}} & Forward Pass + Middle Layer Gradient Compute & 1.507\\
    \midrule
    \ours (\textbf{Ours}) & Forward Pass + Last Layer Gradient Compute & 1.000 \\
    \bottomrule
    \end{tabular}}
    }
    \vspace{.5em}
    \caption{\textbf{Comparison of computational costs.}
    \ours incurst additional computational cost compared to forward-only baselines (\eg, Self-Sup), but only by approximately 3.4\%.
    }
    % \vspace{-1em}
    \label{tab:comp_cost}
\end{table*}

\subsection{Distribution of data selected by selection baselines}
\label{sec:appx:selection_distribution}
The selection baselines that consistently underperform random selection across all selection ratios, namely Self-Sup, COINCIDE, DBP, TIVE, and Adapt-$\infty$, rely on K-means clustering, which often yields imbalanced clusters (\ie, certain clusters contain disproportionately more data) under imbalanced distributions. 
Note that in continual learning, data arrives in a streaming manner, making the temporal distribution at each time point likely imbalanced, even if the overall dataset is balanced across tasks. 
As a result, these methods often lead to skewed data selection, over-sampling data from certain tasks while under-sampling others. 
To quantify this skewness, we measure the difference in the number of selected samples between the most-sampled and least-sampled tasks for each baseline, and summarize the results in Tab.~\ref{tab:selection_dist}. 
As shown, these baselines exhibit significantly higher skewness than our proposed \ours, even higher than random selection, ultimately contributing to their lower performance. In contrast, both \ours and DivBS, the best-performing baseline, achieve much lower selection skewness, indicating a more balanced sample selection across tasks.

\begin{table*}[ht!]
\centering
\begin{tabular}{lcc}
\hline
\textbf{Method} & \textbf{K-means based selection} & \textbf{Max–Min \# of selected samples across tasks} \\
\hline
Self-Sup        & O & 658 \\
COINCIDE        & O & 659 \\
DBP             & O & 687 \\
TIVE            & O & 929 \\
Adapt-$\infty$  & O & 746 \\
Random          & X & 600 \\
DivBS           & X & 520 \\
\ours (Ours)    & X & 482 \\
\hline
\end{tabular}

\caption{\textbf{Comparison of data selection distribution.} Comparison of methods with respect to sample variability across tasks.}
\label{tab:selection_dist}
\end{table*}

\subsection{Comparison of the Number of Selected Samples}
\label{sec:appx:num_samples}
While baselines select a fixed number of samples per batch, thus selecting the same number of total samples, \ours dynamically selects samples probabilistically. 
For fair comparison, we ensure our approach uses comparable or fewer total samples than those selected by other baselines. 
We summarize the selected samples for each baseline in Tab.~\ref{tab:num_selected_samples}. 
Despite using fewer samples, our method outperforms the baselines, as demonstrated in Sec.~\ref{sec:quanti_main}.

\begin{table*}[ht!]
  \centering
  \resizebox{\linewidth}{!}{
    \begin{tabular}{lccccccccc}
    \toprule
    \multirow{2}{*}{Method} & \multicolumn{3}{c}{\oursbenchmark} & \multicolumn{3}{c}{COAST} & \multicolumn{3}{c}{Adapt} \\
   
    & $6.25\%$ & $12.5\%$  & $25.0\%$ &  $6.25\%$ & $12.5\%$  & $25.0\%$ &  $6.25\%$ & $12.5\%$  & $25.0\%$ \\
    
    \cmidrule(lr){1-1} 
    \cmidrule(lr){2-4} \cmidrule(lr){5-7} \cmidrule(lr){8-10}

    Baselines &  43394 & 86788 & 173576 & 10000 & 20000 & 40000 & 40000 &  80000 & 160000 \\
    
    \ours (Ours) & \textbf{43158} & \textbf{86251} & \textbf{173242} & \textbf{9874} & \textbf{19772} & \textbf{39427} & \textbf{39589} & \textbf{79604} & \textbf{159154} \\

    \bottomrule
    \end{tabular}
  }
  \caption{
    \textbf{Comparison of the number of selected samples.} 
    We compare the average number of selected samples by \ours across three different seeds with those selected by other sample selection baselines.
  }
  \label{tab:num_selected_samples}
  \vspace{-.6em}
\end{table*}

\subsection{Detailed Algorithm of \ours}
\label{sec:appx:pseudocode}
We provide a comprehensive pseudocode of \ours in Algorithm~\ref{algo:pseudo}.

\begin{algorithm*}[t!]
\caption{\ours}
\label{algo:pseudo}
\begin{algorithmic}[1]
\State \textbf{Input:} model $f_{\theta}$, batch $\mathcal{B}_t$, batch size $N_\mathcal{B}$, EMA $\mu_t$, EMV $\sigma_t$, threshold $I_T$, number of layers $L$
\State \textbf{Initialize:} Informativeness set $I^{(t)} \leftarrow \emptyset$, high informative sample set $H \leftarrow \emptyset$

\vspace{0.3em}
\State \textbf{// Stage 1: Calculate Informativeness $I^{(t)}$ I of each sample in $\mathcal{B}_t$}
\For{each $(x_i^{(t)}, y_i^{(t)}) \in \mathcal{B}_t$}
    \State Calcuate gradient $g_i \leftarrow \frac{\partial \ell(d^{(t)}_i)}{\partial \theta_L}$
    \State Calculate Informativeness $I_i^{(t)} \leftarrow \text{tr}(g_i \cdot g_i^\intercal)$
    \State Add to set $I^{(t)} \leftarrow I^{(t)} \cup \{I_i^{(t)} \}$
\EndFor

\vspace{0.3em}
\State \textbf{// Stage 2: \ourstwo (\ourstwofull)}
% \State Set temporary informativeness $I^{0} \leftarrow I$
\State Initialize adjusted Informativeness $\widetilde{I}^{(t)} \leftarrow I^{(t)}$
\While{$|H| < N_\mathcal{B}$}
    % \State Descending order sort $\widetilde{I}_i^{(t)}$ for $d^{(t)}_i \in B_t \setminus H$
    % \State Add most informative sample $H \leftarrow H \cup \{d_{|H|}^{(t)}\}$
    \State Add most informative sample $H \leftarrow H \cup \left\{ \arg\max_{d^{(t)}_i \in \mathcal{B}_t \setminus H} \widetilde{I}^{(t)}_i \right\}$
    \For{each $d^{(t)}_i \in \mathcal{B}_t \setminus H$}
        \For{each $d_h^{(t)} \in H$}
            \State Calculate updated Informativeness $\widetilde{I}_i^{(t)} \leftarrow I_i^{(t)} - \text{cos}(g_i, g_h) \cdot I_h^{(t)}$
        \EndFor
            \If{$|H|>1$}
                \State Account higher-order redundancy due to overlapping similarities between $d_h^{(t)} \in H$
                \State $\widetilde{I}_i^{(t)} = \widetilde{I}_i^{(t)} + \sum_{\substack{U \subseteq H \\ |U| \geq 2}} (-1)^{|U|} \text{cos}(g_i, \bar{g}_U) \cdot \bar{I}_{U}^{(t)}$
            \EndIf
    \EndFor
\EndWhile

\vspace{0.3em}
\State \textbf{// Stage 3: Calculate Relative Informativeness $\hat{I}^{(t)}$ }
\For{each $\widetilde{I}_i^{(t)}$ in $\widetilde{I}^{(t)}$}
    \State Calculate Relative Informativeness $\hat{I}_i^{(t)} \leftarrow \frac{\widetilde{I}_i^{(t)} - \mu_t}{\sigma_t}$
\EndFor
\State \textbf{// Stage 4: Probablistic Sampling}
\State Sample random threshold from uniform distribution $r \sim \mathcal{U}(0,1)$
\State Select samples $\mathcal{B}^*_t \leftarrow \{(x_i^{(t)}, y_i^{(t)}) \in \mathcal{B}_t \mid \sigma(\hat{I}_i^{(t)} - I_{T}^{(t)}) > r\}$

\vspace{0.3em}
\State Calculate average Informativeness $\bar{I}^{(t)} = \frac{1}{N_\mathcal{B}}\sum_{i=1}^{N_\mathcal{B}} I^{(t)}_i$
\State Update EMA $\mu_{t} = \beta \bar{I}^{(t)} + (1-\beta)\mu_{t-1}$
\State Update EMV $\sigma_{t} = \beta (\bar{I}^{(t)} - \mu_{t-1})^2 + (1-\beta)\sigma_{t-1} $
\State \textbf{Output:} selected samples $\mathcal{B}^*_t$
\end{algorithmic}
\end{algorithm*}

\subsection{Impact Statements}
\label{sec:appx:impact}

In this work, we focus on advancing an online sample selection algorithm for continual instruction tuning. 
Our approach prioritizes training efficiency while maintaining strong performance, allowing large pre-trained MLLMs to adapt more effectively and make fairer, less biased decisions when processing continuous streams of real-world data.

\subsection{Data Privacy and Content Sensitivity}
For CIT benchmarks, we use existing datasets such as COAST and Adapt, which have already filtered out privacy-sensitive content, as well as \oursbenchmark, a combination of existing multimodal benchmarks that are also free from sensitive content.

\subsection{Potential Risks}
In CIT setup, real-time data arrives continuously in a streaming manner, leading to imbalanced data distributions at each time step. 
This can unintentionally introduce bias throughout training.

\subsection{Parameters for Packages}
For evaluation, we measure accuracy, which selects an answer among multiple candidate choices.
In \oursone, we normalize the informativeness score $I$ using Z-score normalization based on our empirical evidence from the QQ-plot.

\subsection{License For Artifacts}
We utilize publicly available data, models, and codebases, as provided by the original papers for each baseline method.

\subsection{Use of AI Assistance}
We use AI assistance, such as GPT-4, solely for grammatical error corrections.

\end{document}